\documentclass[11pt, a4paper, logo, nonumbering]{style}
\usepackage[numbers, sort&compress]{natbib}
\usepackage{dblfloatfix}
\usepackage{caption}
\usepackage{xspace}
\usepackage{pifont}
\usepackage{multirow}
\usepackage{tcolorbox}
\usepackage{longtable}
\usepackage{hyperref}
\usepackage{amsfonts}
\usepackage{amsmath}
\usepackage{amssymb}
\usepackage{adjustbox}

\usepackage[bottom]{footmisc}
\usepackage{listings}
\usepackage{CJKutf8}
\usepackage{setspace}

\usepackage{array}
\usepackage{tabularx}
\usepackage{booktabs}
\usepackage{colortbl}

\usepackage{multicol}

\usepackage{booktabs}
\usepackage{tabularx}
\usepackage{array}
\usepackage{longtable}

\newcolumntype{Y}{>{\raggedright\arraybackslash}X}
\newcolumntype{L}[1]{>{\raggedright\arraybackslash}p{#1}}

\usepackage{pgfplots}
\pgfplotsset{compat=1.18}
\definecolor{oursblue}{HTML}{2C6FB5}
\definecolor{basegray}{HTML}{7A7F87}
\definecolor{goodgreen}{HTML}{1B7837}
\definecolor{badred}{HTML}{B2182B}

\definecolor{codebackground}{rgb}{0.95,0.95,0.95}
\definecolor{codeframe}{rgb}{0.8,0.8,0.8}
\definecolor{keyword}{rgb}{0.0,0.0,0.55}
\definecolor{stringcolor}{rgb}{0.65,0.13,0.13}
\definecolor{commentcolor}{rgb}{0.0,0.5,0.0}

\lstdefinestyle{code}{
    backgroundcolor=\color{codebackground},
    frame=single,
    rulecolor=\color{codeframe},
    basicstyle=\ttfamily\footnotesize,
    keywordstyle=\bfseries,
    stringstyle=\color{stringcolor},
    commentstyle=\itshape\color{commentcolor},
    showspaces=false,
    showstringspaces=false,
    showtabs=false,
    tabsize=4,
    captionpos=b,
    breaklines=true,
    breakatwhitespace=true,
    sensitive=true,
    breakindent=0pt,
}

\makeatletter
\def\@BTrule[#1]{%
  \ifx\longtable\undefined
    \let\@BTswitch\@BTnormal
  \else\ifx\hline\LT@hline
    \nobreak
    \let\@BTswitch\@BLTrule
  \else
     \let\@BTswitch\@BTnormal
  \fi\fi
  \global\@thisrulewidth=#1\relax
  \ifnum\@thisruleclass=\tw@\vskip\@aboverulesep\else
  \ifnum\@lastruleclass=\z@\vskip\@aboverulesep\else
  \ifnum\@lastruleclass=\@ne\vskip\doublerulesep\fi\fi\fi
  \@BTswitch}
\makeatother

\addto\extrasenglish{
}

\renewcommand{\today}{}
\renewcommand{\thefootnote}{\fnsymbol{footnote}}
\makeatletter
\newcommand\blfootnote[1]{%
  \begingroup
  \renewcommand\thefootnote{}\footnote{#1}%
  \addtocounter{footnote}{-1}%
  \endgroup
}
\makeatother

\definecolor{absframe}{HTML}{ED6000}
\makeatletter
\renewcommand{\abscontent}{%
  \noindent
  \begin{tcolorbox}[colframe=absframe, colback=absframe!5, boxrule=1pt, arc=8pt,
                    left=10pt, right=10pt, top=10pt, bottom=10pt]
    \centerline{\fontsize{15pt}{14pt}\selectfont\textbf{Abstract}}\vspace{2ex}
    {\absfont \theabstract}%
    \@ifundefined{@keywords}{}{%
      \vskip1em \noindent \keywordsfont Keywords: \@keywords}%
  \end{tcolorbox}%
}
\makeatother

\title{\centering Agentic Routing: The Harness-Native Data Flywheel}

\author[*]{
TokenRhythm Technologies
\\
\small
\href{https://github.com/opensquilla/opensquilla}{\textcolor{absframe}{\texttt{https://github.com/opensquilla/opensquilla}}}
}

\newcommand{\opensquilla}{\textsc{OpenSquilla}\xspace}

\newcommand{\harnessnative}{\textsc{Harness-Native}\xspace}

\begin{abstract}
Large language model agents are increasingly executed not by a single model call, but by an execution harness that manages observation, context, control, action, state, and verification. At the same time, frontier and open models are becoming structurally specialized: a model that is strong at code editing, long-context recovery, tool use, mathematical reasoning, or low-latency response may not dominate on the other axes. This makes model selection inside an agent a core systems problem rather than a per-query serving trick. Existing routing methods mostly optimize single-turn cost-quality trade-offs and therefore miss the execution state, intermediate failures, and feedback loops that make agents different from chat completion. We propose \harnessnative agentic routing, a step-level routing paradigm that selects either a single best-fit model for cost-effective execution or multiple complementary models for ensemble-style accuracy improvement, conditioned on the full harness state. The key insight is that every routing decision naturally produces a structured data record---consisting of the query, harness state, model choice or model set, execution trace, outcome, and cost---whose labels are supplied by the environment rather than by the router itself. These records form a \emph{harness-native data flywheel}: execution traces train better routers and harness-native models, which improve cost-quality trade-offs and generate more traces under the same budget. We instantiate this idea in \opensquilla with a four-layer routing stack, an open LightGBM cold-start ranker, and a staged router-model path that turns logged arena records into progressively stronger routing policies. The report studies singleton and multi-model routing on agentic benchmarks including DRACO and PinchBench, and argues that agentic routing is not merely cost control, but a data engine for agent-native training.
\end{abstract}

\begin{document}
\begin{CJK*}{UTF8}{gbsn}
\maketitle

\section{Introduction}\label{sec:intro}

An AI agent is traditionally defined by a simple decomposition: \emph{Agent = Model + Harness}~\cite{langchain2026harness}, where a single foundation model supplies the intelligence and the harness wraps it into a working system. As base models proliferate and become increasingly specialized---in coding, long-context recovery, reasoning, tool use, or low-latency response---no single model remains the best choice at every step, and the recent evolution of AI agents is better summarized by an updated decomposition: \emph{Agent = Base Models + Harness}~\cite{guo2026question}. Base models~\cite{anthropic2026claudeopus47,singh2025openai} provide latent capabilities such as language understanding, reasoning, code generation, tool-use priors, and multimodal perception, while the harness~\cite{steinberger2026openclaw,opensquilla2026,nousresearch2026hermesagent} provides the execution substrate that turns these capabilities into observable, controllable, and verifiable behavior. The two sides are complementary rather than hierarchical: a model without a harness remains a passive generator, and a harness without capable base models has no intelligence to orchestrate. Progress in agents is therefore an alternating process rather than a one-sided model-scaling story. Stronger base models make new behaviors possible; harness engineering converts those behaviors into reliable workflows; those workflows expose new bottlenecks such as context drift, brittle tool calls, poor recovery, or unverifiable intermediate states; and these bottlenecks in turn define what the next generation of models should internalize. In this view, the transition from prompt engineering to harness engineering and agent-native training is a co-evolution between base models and the execution systems that make them useful.

At the same time, models are becoming increasingly specialized rather than interchangeable. Contemporary model families differ not only in average benchmark scores but also in their operational profiles: coding-oriented models may be more competitive on repository-scale editing, long-context models may be stronger at evidence recovery, reasoning models may be better for mathematics but slower and more expensive, lightweight models may be ideal for routine control decisions, and tool-use models may tolerate different action formats. These differences create a Pareto frontier across quality, latency, cost, context length, tool reliability, and safety instead of a single model that dominates every axis. The specialization is structural because many objectives compete with one another: fast and slow thinking favor different inference regimes, concise tool calls and deep reasoning pull output distributions in different directions, and a model's apparent ability often changes with the harness in which it is embedded.

\begin{figure}[t]
  \centering
  \includegraphics[width=1.0\linewidth]{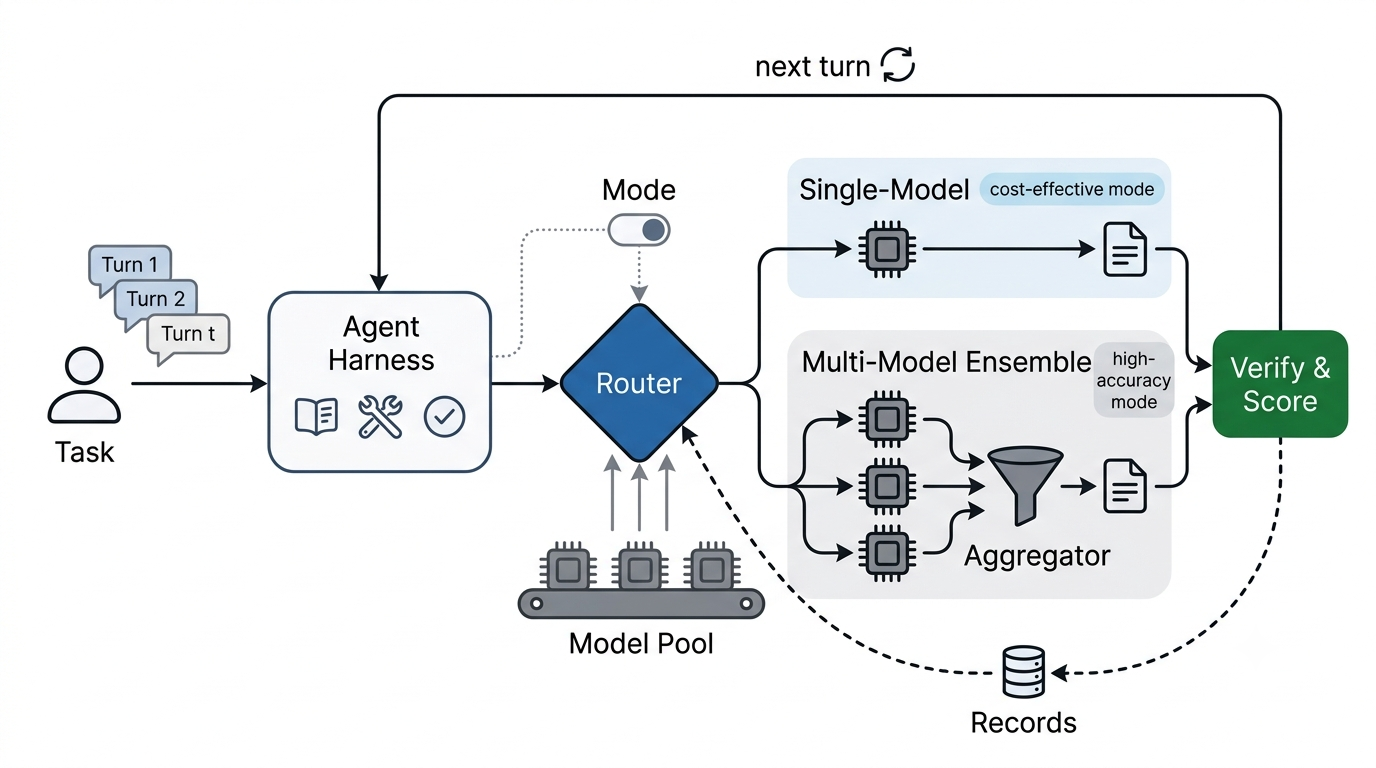}
  \caption{\textbf{Overview of the two operating regimes of agentic routing.}
The agent harness supplies the execution state to the router, which selects from a shared model pool. The user-selected mode determines the regime: in the cost-effective mode the router fields a single best-fit model, while in the high-accuracy mode it fields several complementary proposers whose outputs are fused by an aggregator. Verification and cost signals score each decision, and the resulting records feed back to improve the router.}
  \label{fig:single-multi-router}
\end{figure}

When model specialization is combined with the large adaptation space of the harness, routing is no longer the problem of simply choosing which model should answer a query. It becomes a joint optimization problem over a specialized model space and a specialized harness adaptation space. A routing decision can determine not only the model, but also how observations are formatted, how context is compressed, whether the control loop should use a cheap fast path or a stronger deliberative path, which action schema is exposed, how state is carried across steps, and what verification or recovery policy is applied after a failure. Model routing should be treated as a meta-decision across harness responsibilities rather than as a thin serving layer above fixed prompts.

We call this paradigm \harnessnative agentic routing: step-level model selection during agent execution conditioned on the full harness state. At each decision point, the router can operate in two regimes, selected by the user as a deployment mode (Figure~\ref{fig:single-multi-router}). In the cost-effective regime, it selects a single best-fit model from the current model pool, and potentially a harness configuration, to achieve the best cost-quality trade-off for the current dialogue turn or execution step. In the high-accuracy regime, it selects a set of complementary models and aggregates their outputs through ensemble, verification, voting, debate, or fallback policies to improve task accuracy beyond any single cheap model. This formulation differs from query-level model routing because agent execution is multi-step, stateful, and recoverable: a cheap decision at one step may force costly correction later, while a strong model or a multi-model ensemble may be unnecessary for routine control but essential after verification detects drift. The router must therefore reason over execution trajectories, not only prompts, and it must treat cost, latency, and quality as coupled Pareto dimensions. Crucially, this definition also makes routing a data-producing operation. Every step-level decision can be logged as a structured record containing the query, the harness state, the model choice or model set, the execution trace, the outcome, and the realized cost. Unlike preference data or static benchmark labels, these records are produced by the agent's own execution environment: the model choice is an action, while the outcome and cost are supplied by verification and runtime accounting as ground-truth labels. Routing decisions are therefore not only control decisions, but also the starting point of a harness-native data flywheel.

The resulting data flywheel closes the loop between routing, harness execution, and model training. Better routing increases the fraction of steps handled by cheaper capable models, which lowers cost under a fixed budget. Lower cost allows the harness to execute more tasks and collect more traces. More traces improve router models and create supervision for harness-native specialist models. These specialist models are then injected back into the model pool rather than replacing general foundation models, making the pool more diverse and increasing the value of routing. The positive feedback is not an assumption that the router is already good; it comes from environment-labeled traces, exploration coverage, and the label variance provided by a strong base-model pool.

We instantiate this idea in \opensquilla for both singleton routing and multi-model ensemble routing. The open router follows a four-layer design: a token-complexity filter sends trivial requests to cheap models, a task-type classifier separates coding, reasoning, chat, and tool-use cases, a context-aware refiner incorporates harness state, and a LightGBM ranker selects the cheapest capable model from the remaining candidates. We also define a staged router-model path for two deployment regimes: a cost-saving single-model mode that reduces cost at similar quality, and a high-accuracy multi-model mode that selects complementary models from a richer pool and combines their outputs to improve end-to-end quality beyond single-model routing. The open-source boundary is deliberate: the LightGBM router can be released, while accumulated arena records and later high-accuracy policy generations can be served through the \opensquilla API.

Our evaluation covers the two deployment regimes of agentic routing. For singleton routing, we measure whether a routed policy can preserve end-to-end task quality while reducing realized cost on PinchBench and DRACO. For multi-model routing, we measure whether selected complementary proposer sets and aggregation improve the cost--quality operating point on DRACO and PinchBench. The evaluation reports aggregate deployed-policy frontier points, while the logged arena records provide the basis for route-level decomposition of routing granularity, harness-state signals, and model-pool effects.

Aggregate results show that the open LightGBM router can preserve task quality while cutting realized cost by roughly $90\%$ on PinchBench and about $43\%$ on DRACO relative to a fixed strong-model baseline, and that multi-model routing can push the DRACO quality frontier upward while often lowering cost.

\section{Related Work}\label{sec:related}

\paragraph{Model routing.}
Model routing studies how to improve the cost-quality trade-off of LLM serving by selecting among models with different prices and capabilities. Prior systems use cascades \citep{chen2023frugalgpt}, difficulty or meta-model estimation \citep{sakota2024flyswat,ding2024hybrid}, preference data \citep{ong2024routellm}, and model-specific performance or benchmark-derived outcomes \citep{mohammadshahi2024routoo,shnitzer2023large} to decide whether a request can be handled by a cheaper model or should be escalated to a stronger one. Recent benchmarks standardize this setting with multi-model outcome matrices \citep{hu2024routerbench,li2026llmrouterbench}, while deployment-oriented work studies budget constraints, user-controlled cost-quality tolerance, and lightweight or training-free routing \citep{panda2025adaptive,feng2025ipr,chen2025tagrouter}. A related line of work ensembles multiple LLM outputs through ranking, fusion, or multi-agent aggregation to improve quality \citep{jiang2023llmblender,wang2025mixture}. These works establish that many inputs do not require the strongest model, but their routing problem is usually defined over a static prompt or a single input-output pair. The router observes the user request, and sometimes a predicted quality score, but not the execution state of an agent. It therefore does not directly model tool failures, context pressure, verifier feedback, recovery attempts, or the accumulated trajectory that determines whether a cheap model remains sufficient at the current step.

\paragraph{Agent harnesses and agent-native data.}
Recent agent systems show that model behavior is strongly shaped by the surrounding harness, including tool schemas, observation interfaces, memory, context management, verification loops, and recovery policies \citep{guo2026question,yao2023react,shinn2023reflexion,wang2024voyager,yang2024sweagent}. Work on tool use and agent tuning further suggests that agent data should encode actions, feedback, and task execution rather than only static instruction-response pairs \citep{schick2023toolformer,chen2024agentflan,zeng2024agenttuning}. Some recent routing methods have begun to move toward agentic settings, such as routing in multi-agent collaboration, tool calling, explainable agent workflows, or preference-aligned domain-action selection \citep{yue2025masrouter,agarwal2026switchcraft,okamoto2026explainable,tran2025archrouter}. Our work connects these threads by treating routing not only as a serving-time optimization, but also as a harness-native data interface. At each agent step, the routing decision reflects the current task, harness state, available skills or tools, verification signals, recovery history, and the cost and risk of alternative model choices. By recording which model or model set was sufficient under a given trajectory, and when escalation, verification, or recovery was needed, agentic routing produces structured supervision for future agent-native training.

\section{What is Agentic Routing}\label{sec:method}

A traditional agent instantiates the decomposition \emph{Agent = Model + Harness}: one foundation model $\bar{m}$ is fixed offline, every harness step calls the same model, and all optimization happens on the harness side---prompts, tools, context management, and recovery policies are tuned around the fixed model. In the language of this section, such an agent has no routing operator: the model choice is a constant, the realized cost profile is dictated by $\bar{m}$, and the only way to trade quality against cost is to swap the entire agent onto a different model. Once the pool $\mathcal{M}$ contains many base models, this constant becomes a per-step decision variable. Agentic routing replaces the fixed $\bar{m}$ with a state-conditioned routing operator $g$ that decides, at every execution step, which model or model set should act next, turning the quality--cost trade-off into an explicit optimization target; the traditional agent is recovered as the degenerate router $g\equiv\{\bar{m}\}$. This section formalizes the operator and its objective.

\subsection{Problem Definition and Objective}\label{sec:definition}

Agentic routing is step-level model or model-set selection during agent execution conditioned on the harness state. Let $q$ be a user task, $\mathcal{M}$ be a pool of candidate models, and $h_t$ be the harness state at decision step $t$. The state $h_t$ includes the current observation, compressed and raw context, control-loop status, available actions, artifact state, tool history, recovery status, and verification signals.

\begin{figure}[t]
  \centering
  \includegraphics[width=1.0\linewidth]{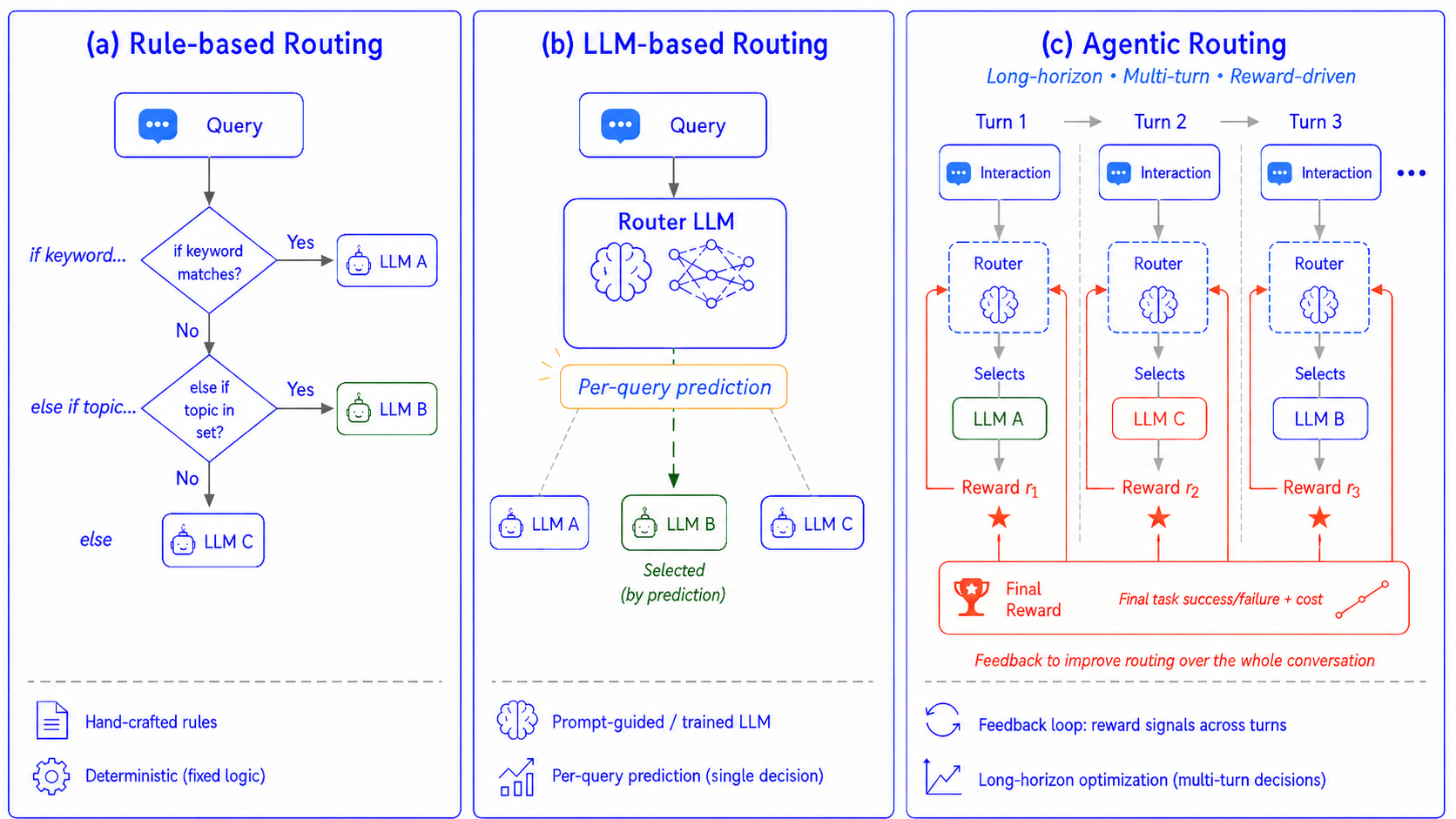}
  \caption{\textbf{Architectural comparison of different model routing mechanisms.}
(a) \textbf{Rule-based Routing}: A static approach that relies on hard-coded heuristic rules and keyword-based intent classification to assign user queries to specific models.
(b) \textbf{LLM-based Routing}: A semantic-aware routing mechanism where a LLM dynamically serves as the router to make the optimal model selection.
(c) \textbf{Agentic Routing (ours)}: A step-level harness-native router observes execution state, tool and verifier signals, recovery status, and route history, then selects a model or model set for the next harness step. The resulting outcome, cost, and verification signals are logged as arena records for later router improvement.}
  \label{fig:agentic_routing}
\end{figure}

We do not define routing as a single cost-minimization problem. Prior model-routing and cascading systems have shown that heterogeneous model pools can improve the cost-quality trade-off of LLM serving, but they usually operate at the level of static prompts or queries~\citep{chen2023frugalgpt,ong2024routellm,hu2024routerbench,moslem2026routingsurvey}. Recent agentic routing benchmarks and routers move this decision closer to execution trajectories~\citep{yang2026twinrouterbench,zhou2026agentasarouter}. Figure~\ref{fig:agentic_routing} contrasts rule-based and LLM-based routing with this harness-native view. Our formulation takes the next step: a router may preserve quality at lower cost, compose cheaper specialist models to dominate a strong singleton baseline, or spend more budget at high-risk states to improve task success. We therefore define a single quality-cost frontier for harness-native agent execution.

We model routing as an operator $g$ that, conditioned on the current harness state $h_t$, selects a set of models $S_t=g(\mathcal{M}\mid h_t)$ from the pool $\mathcal{M}$ for the next execution step and, when it selects more than one, a policy for combining them. The router optimizes the Pareto front between task loss and cost:
\begin{equation}
\min_{g}\;
\Big(\;
\mathbb{E}_{q\sim\mathcal{D},\,\tau\sim g}\big[\ell(\tau)\big],
\;\;
\mathbb{E}_{q\sim\mathcal{D},\,\tau\sim g}\big[C(\tau)\big]
\;\Big),
\qquad
S_t=g(\mathcal{M}\mid h_t),
\quad
1 \le |S_t| = k \le K,
\label{eq:agentic-routing-frontier}
\end{equation}
where the objective pair is minimized in the Pareto sense, $\ell(\tau)=1-R_{\mathrm{task}}(\tau)$ is the task loss of the resulting trajectory, $C(\tau)=\sum_{t=1}^{T}\sum_{m\in S_t} c(m,h_t)$ is its realized cost, and $k=|S_t|$ is the number of models fielded at step $t$, capped by $K$. Setting $k=1$ recovers singleton routing (Section~\ref{sec:singleton-routing}), where $g$ fields exactly one model; $k>1$ gives multi-model routing (Section~\ref{sec:multi-model-routing}), where $g$ fields several models together with a policy for combining them. Equivalently, sweeping the scalarized objective $\mathbb{E}_{\tau\sim g}[\ell(\tau)+\lambda\,C(\tau)]$ over $\lambda\ge 0$ traces the same frontier, and latency enters as an optional third axis when a deployment requires it.

Reading Equation~\ref{eq:agentic-routing-frontier}, a router $g$ is better when it lowers task loss, lowers cost, or both. Crucially, a multi-model action ($k>1$) does not necessarily cost more than a singleton action, because $S_t$ may combine several cheap or specialist models whose total cost is below that of a single expensive frontier model. Different operating points on the frontier answer different questions: a low-cost point asks how cheaply the harness can preserve task quality, while a high-accuracy point spends more budget at difficult, high-risk, or weakly recoverable states. We instantiate this frontier along the model and model-set dimension of the routing action: the action fields a model set $S_t$, together with an aggregation policy when $k>1$. To isolate the model-allocation component of the routing action, this work fixes the surrounding harness policies and evaluates how far the frontier can move through model and model-set selection alone. This makes the reported gains attributable to capability allocation under a fixed execution substrate.

The loss is task-level rather than purely step-level. Agent execution is recoverable: a locally weak step can be corrected later, while a locally cheap step can create downstream recovery cost. This view is consistent with agent work that treats reasoning, acting, tool use, memory, and feedback as interleaved execution rather than isolated text generation~\citep{yao2022react,schick2023toolformer,shinn2023reflexion,wang2023voyager,yang2024sweagent}. The supervision signal therefore should not assume that every step has an independent correctness label; the natural reward is terminal or verification-derived, and $g$ must learn from trajectories rather than isolated prompts. The trajectory fields
\begin{equation}
(q,\;h_t,\;S_t,\;z_t,\;c_t,\;y),
\end{equation}
define the schema of harness-native arena records: execution records annotated with outcome, cost, and verification signals. In this report, the open LightGBM router is the cold-start policy that starts this record stream, while later router-model iterations use the accumulated arena records to improve future routing decisions.

\subsection{Singleton Routing Model}\label{sec:singleton-routing}

We first instantiate the frontier in the singleton setting. Setting $k=1$ in Equation~\ref{eq:agentic-routing-frontier} restricts the routing operation to a single model,
\begin{equation}
S_t=g(\mathcal{M}\mid h_t)=\{m_t\}.
\end{equation}
The router still optimizes the same task-level frontier, but its online action is now to field exactly one model from a heterogeneous model pool for the next harness step.

We call this view \emph{capability matching}. The harness issues a state-conditioned capability demand; the model pool supplies heterogeneous capability profiles with different prices, latencies, context limits, tool-use reliability, and failure modes; and the singleton router selects the cheapest risk-adjusted model whose capability is sufficient for the current state. The model pool is thus an arena of specialists, and the router pairs each state's demand with the model best suited to it under uncertainty. The router is not merely classifying the prompt. It is allocating model capability inside the execution economy of the harness, where a superficially cheap model can become expensive if its failure causes retries, tool repair, context reconstruction, or downstream verification recovery.

Let the singleton routing state be
\begin{equation}
x_t=(q,h_t),
\end{equation}
where $q$ is the original user task and $h_t$ is the current harness state. The harness state includes not only the visible prompt, but also context pressure, artifact state, tool history, recovery status, recent routing decisions, and verification signals. The required capability is therefore not a fixed property of the user query. It is a trajectory-dependent demand induced by the current execution state.

Operationally, singleton routing is Equation~\ref{eq:agentic-routing-frontier} restricted to $k=1$ and applied greedily at the current step. To make this per-step decision easier to estimate and optimize than a purely terminal signal, we add a \emph{per-turn reward} $\rho_t=\rho(h_t,u_t,z_t)\in[0,1]$ that scores the immediate outcome $z_t$ of the routing action: a verifiable reward such as a passing unit test, a successful tool call, or a satisfied schema or safety constraint, or an LLM-as-judge score of the step output. Fixing the trade-off weight $\lambda$ of Equation~\ref{eq:agentic-routing-frontier}, the router fields the single model that minimizes the expected task loss and cost for the remainder of the trajectory, discounted by this immediate reward:
\begin{equation}
m_t
=
\arg\min_{m\in\mathcal{M}_t}\;
\mathbb{E}_{\tau\sim g}\!\left[\,
\ell(\tau) + \lambda\,C(\tau) - \beta\,\rho_t
\;\middle|\;
S_t=\{m\},\, h_t
\,\right].
\label{eq:singleton-clearing}
\end{equation}
The weight $\beta\ge 0$ trades the immediate per-turn reward against the trajectory-level task loss $\ell(\tau)$ and cost $C(\tau)$; setting $\beta=0$ recovers the purely terminal objective of Equation~\ref{eq:agentic-routing-frontier}. Keeping $\ell$ and $C$ trajectory-level preserves the recoverable-execution semantics---a superficially cheap model that later triggers retries, tool repair, context reconstruction, or verification recovery still raises the downstream cost $C(\tau)$, while a model that under-serves the current state raises the task loss $\ell(\tau)$---so the per-turn reward only \emph{densifies} the supervision, giving the router immediate step-level feedback that eases credit assignment and speeds optimization. The capability-matching intuition---matching a state-conditioned capability demand against the heterogeneous capability supply of the pool, and pricing residual risk for irreversible or weakly recoverable actions---is a reading of how $\ell(\tau)$, $C(\tau)$, and $\rho_t$ respond to the choice of $m$, not a separate optimization problem. The main point is the same: a model is not cheap if its failure creates expensive recovery, and a model is not strong if its capability does not match the state-specific demand.

The required capability profile is multi-dimensional. Routine control, short tool-call formatting, low-risk context bookkeeping, or cacheable response generation may require only a lightweight model. Ambiguous planning, difficult code edits, long-context debugging, recovery after verification failure, safety-sensitive decisions, or final artifact generation may require a stronger model. The same user task can therefore require different models at different steps. Context pressure can make long-context reliability more important than raw reasoning score; a recent tool error can make action-format robustness more important than benchmark performance; a failed verifier can raise the required capability for the next recovery step. This is the difference between harness-native singleton routing and ordinary query-level routing: the model choice is conditioned not only on what the user asked, but on where the harness currently is in the execution trajectory.

The open singleton router in \opensquilla implements this matching view as a lightweight cold-start mechanism. It performs four passes: \emph{order admission}, which sends clearly trivial and low-risk states to cheap fast paths; \emph{demand construction}, which builds a coarse capability profile from the task and harness state; \emph{risk pricing}, which adjusts that demand using context pressure, tool failures, verifier results, recovery status, action irreversibility, and downstream correction risk; and \emph{capability matching}, which binds the priced demand to a concrete model or tier through the deployment registry. This decomposition is deliberately stronger than a model-ranking pipeline. It separates the question of what the next harness step requires from the question of which currently deployed model should satisfy it.

In the released open implementation, the first three passes are realized by cheap deterministic filters and lightweight classifiers, while the final matching pass is implemented by a LightGBM ranker over the remaining model  choices~\citep{ke2017lightgbm}. LightGBM is therefore useful but not fundamental. It is the cold-start matching engine, not the market itself. The market consists of the harness state, the model pool, the deployment registry, the verifier, and runtime accounting. The harder learning problem is to infer, from harness-native trajectories, the state-conditioned capability demand, the risk of under-routing, and the expected recovery cost that Equation~\ref{eq:singleton-clearing} prices only implicitly through $\ell(\tau)$ and $C(\tau)$.

This positioning is important for the open-source boundary. The LightGBM router is the first practical point on the singleton frontier: sample-efficient enough for cold start, fast enough for step-level routing, inspectable enough for open release, and simple enough to improve from logged execution. It should not be read as the final router model. It is the open seed that starts the arena-record stream.

The router model should instead be viewed as an \emph{iterated matching policy}. Each batch of arena records updates the same latent objects that define singleton routing: demand construction, risk pricing, and capability matching. The state and selected model are logged as the action; final outcome, verifier events, recovery cost, realized cost, and latency are supplied by the environment. These records allow later router-model iterations to learn when the open policy under-routes a state, when it overpays for unnecessary capability, and when a model's deployment profile has shifted because the model pool or harness has changed.

In the singleton setting, router-model iterations are evaluated in cost-saving mode: each iteration must field one model for the current step while preserving task quality and reducing unnecessary calls to expensive models. The learning target is outcome improvement rather than imitation of the previous router, matching the logged-feedback and off-policy learning view in contextual bandit evaluation~\citep{dudik2011doublyrobust}. A bad LightGBM decision is therefore useful feedback rather than poisoned supervision, because the environment supplies the label through verification, recovery, realized cost, and latency.

In this sense, singleton routing is not model selection in the narrow serving-layer sense. It is outcome-aware allocation of model capability under harness-state uncertainty. The LightGBM router is the open seed of this capability-matching flywheel, while the router model under iteration is the next-stage matching policy progressively learned from the records produced by that seed.

\subsection{Multi-Model Ensemble Routing}\label{sec:multi-model-routing}

Multi-model ensemble routing is the $k>1$ instantiation of Equation~\ref{eq:agentic-routing-frontier}: the router $g$ fields a model \emph{set} rather than a single model. Given the harness state $h_t$ and the candidate pool $\mathcal{M}_t \subseteq \mathcal{M}$ that remains after the context-aware refiner prunes $\mathcal{M}$ under $h_t$, the routing operation returns a proposer set together with a combination policy,
\begin{equation}
g(\mathcal{M}_t \mid h_t) = (P_t, a_t),
\qquad
P_t \subseteq \mathcal{M}_t,
\quad
a_t \in \mathcal{G},
\end{equation}
where the proposers in $P_t$ generate candidate outputs and the aggregation policy $a_t$ fuses the candidates into a single executable result; $\mathcal{G}$ contains voting, verifier-based selection, judge models, and fallback policies. When the aggregator is itself a model call, the selected set of Equation~\ref{eq:agentic-routing-frontier} is
\begin{equation}
S_t = P_t \cup \{m_t^{\mathrm{agg}}\},
\qquad
k = |S_t|,
\end{equation}
where $m_t^{\mathrm{agg}}$ is the aggregator model chosen as part of $a_t$; for rule-based, voting, or verifier aggregators, $S_t = P_t$. The cost $C(\tau)$ of Equation~\ref{eq:agentic-routing-frontier} now accounts for every proposer call and the aggregation overhead, while latency is tracked separately as an overhead metric.

As in the singleton case, the same frontier is instantiated per step. Write $\ell(u \mid h_t)=\mathbb{E}_{\tau\sim g}[\ell(\tau)\mid u_t=u,\,h_t]$, $C(u \mid h_t)=\mathbb{E}_{\tau\sim g}[C(\tau)\mid u_t=u,\,h_t]$, and $\rho(u \mid h_t)=\mathbb{E}_{\tau\sim g}[\rho_t\mid u_t=u,\,h_t]$ for the expected task loss, cost, and per-turn reward of a routing action $u$ at state $h_t$. With the multi-model action $u_t=(P_t,a_t)$, the router minimizes the scalarized objective of Equation~\ref{eq:agentic-routing-frontier}, augmented with a complementarity regularizer and the per-turn reward:
\begin{equation}
(P_t,a_t)
=
\arg\min_{P\subseteq\mathcal{M}_t,\ a\in\mathcal{G}}
\;
\Big[\,
\ell\big((P,a)\mid h_t\big)
+
\lambda\,C\big((P,a)\mid h_t\big)
-
\alpha\,V\big(P\mid h_t\big)
-
\beta\,\rho\big((P,a)\mid h_t\big)
\,\Big].
\label{eq:multimodel-clearing}
\end{equation}
The first two terms are the scalarized frontier objective of Equation~\ref{eq:agentic-routing-frontier} at $k>1$, applied greedily at the current step; the third term $-\alpha V(P\mid h_t)$ is a complementarity regularizer with weight $\alpha$, defined below; and the fourth term $-\beta\,\rho((P,a)\mid h_t)$ is the per-turn reward of Equation~\ref{eq:singleton-clearing} extended to the ensemble action, with the verifier or LLM judge applied to the aggregated output. The singleton rule of Equation~\ref{eq:singleton-clearing} is the special case $u_t=(\{m\},\varnothing)$, where $V$ vanishes.

The cost $C$ aggregates the proposer calls and the aggregation overhead, so a larger set is penalized exactly to the extent that it spends more. The complementarity term $V(P\mid h_t)$ rewards proposer sets whose members fail in \emph{decorrelated} ways. We implement $V(P\mid h_t)$ with a diminishing-returns surrogate based on historical error decorrelation, state-conditioned disagreement, and verifier feedback, which supports efficient greedy subset construction in practice. In principle the task loss $\ell$ already prefers such sets, since proposers that reduce correlated failures lower the expected loss while redundant models sharing training distributions and failure modes raise cost without lowering loss. But because $\ell$ is trajectory-level, off-policy, and noisy while the choice of $P$ is combinatorial, the explicit term $-\alpha V$ acts as a regularizer that makes the subset search easier to optimize and generalize, steering exploration toward complementary sets instead of collapsing onto the logging policy's observed favorites. It is an inductive bias on the surrogate, not a change to the frontier of Equation~\ref{eq:agentic-routing-frontier}: $\alpha$ is kept small and annealed toward zero as the loss estimates sharpen, so the true cost--quality frontier is recovered as the estimator improves and the router is never rewarded for diversity that does not reduce loss. This complementarity term is what separates multi-model routing from top-$k$ selection. The size cap $|P|\le K$ is inherited from Equation~\ref{eq:agentic-routing-frontier}, and a latency budget enters, when a deployment requires it, as the optional third axis of Equation~\ref{eq:agentic-routing-frontier}.

The two regimes are two readings of Equation~\ref{eq:multimodel-clearing} at different values of the cost weight $\lambda$, compared against the strong-singleton reference action $u_t^{\mathrm{ref}}=(\{m_t^{\mathrm{ref}}\},\varnothing)$, where $m_t^{\mathrm{ref}}$ is a strong single-model reference for the current state. When $\lambda$ is light, the minimizer favors the accuracy-seeking ensemble: it spends additional calls at states where a single-model error is costly and hard to recover from---final answer generation, complex code modification, irreversible tool actions, safety-sensitive decisions, long-context synthesis, or low router confidence---driving the loss below the reference, $\ell(u_t\mid h_t) < \ell(u_t^{\mathrm{ref}}\mid h_t)$, at the price of higher cost up to a per-step budget $C(u_t\mid h_t)\le B_t$. A typical division of labor lets one model produce the primary proposal, another identify errors from a different perspective, and a third provide a cheap sanity check. When $\lambda$ is heavy, the same minimizer favors the cost-saving ensemble: it composes medium- or low-cost models to replace the expensive reference, keeping the loss within a tolerance $\epsilon$ of the reference, $\ell(u_t\mid h_t)\le\ell(u_t^{\mathrm{ref}}\mid h_t)+\epsilon$, while spending strictly less, $C(u_t\mid h_t)<C(u_t^{\mathrm{ref}}\mid h_t)$; when the loss is also strictly below the reference, the same composition is simultaneously accuracy-improving. Both conditions follow from Equation~\ref{eq:multimodel-clearing} rather than adding new objectives. The economics of multi-model routing is thus a search for a better cost--quality operating point among a strong singleton, a cheap singleton, and a composition of cheaper models.

We implement multi-model routing as a two-stage procedure. Stage one is task-profile-driven subset selection: the routing model first emits a task profile $\phi_t$ computed from $(q,h_t)$, summarizing signals such as task type, difficulty, required capabilities, risk level, verification availability, cost and latency tolerance, and routing uncertainty. This profile is a compact representation of the state-conditioned quantities in the objective above rather than a new environment state. Given $\phi_t$ and the candidate pool $\mathcal{M}_t$, the selector chooses a proposer subset $P_t$ whose capabilities match the profile and whose members provide complementary evidence. Additional proposers are scored by their fit to the task profile, their own cost and latency profiles, expected marginal value, verification compatibility, uncertainty reduction, and complementarity with the models already selected. A valuable proposer is not merely strong in isolation; it supplies a capability or perspective required by the task profile, such as long-context evidence recovery, code synthesis, adversarial critique, or low-cost error detection. Selection is therefore profile-conditioned subset construction rather than top-$k$ ranking, and the preference is regime-dependent: reliability gain at critical states for the accuracy-seeking ensemble, quality retention at lower total cost for the cost-saving ensemble.

Stage two is state-aware aggregation. Each proposer $m \in P_t$ independently generates a candidate $r_m$, and the aggregator produces the final result
\begin{equation}
\hat{r}_t
=
a_t\big(\{(m,r_m)\}_{m\in P_t}, h_t\big).
\end{equation}
The policy $a_t$ is also conditioned on the task profile and the strongest verification signal observable to the harness: executable signals such as unit tests, static analysis, or patch application in software tasks; schema, permission, and safety constraints in tool-use tasks; rubric-based judging, consistency, factuality, and citation checks in open-ended research tasks. $a_t$ is not a post-processing module but part of the routing action: the router jointly decides which models propose and how their candidates are fused. In the cost-saving regime the aggregator must itself be cost-aware, since an expensive aggregator over cheap proposers can eliminate the proposer-side savings; $a_t$ is then a lightweight judge, a rule-based or verifier-based selector, or a low-cost model aggregator. The accuracy-seeking regime may use a stronger aggregator when task risk, verification difficulty, and the resulting quality improvement justify its cost.

Each ensemble step is also a data-producing operation: the per-proposer candidates $r_m$ provide outcomes from several models on the same $(q, h_t)$ decision point, contributing to the counterfactual coverage that Section~\ref{sec:routing-data} builds through controlled exploration. The two-stage design preserves the singleton cost--quality trade-off: the system degenerates to singleton routing in routine states, selectively introduces complementary proposers where a single-model failure is expensive, and replaces a strong singleton with a composition of cheaper models in cost-sensitive states---expanding the quality frontier while reducing the cost of reaching a given quality level.

\subsection{Router-Model Iterations: Implementation and Evolution}\label{sec:router-model-iterations}

The router is deployed not as a single model but as a sequence of policies $g^{(0)}, g^{(1)}, g^{(2)},\dots$ that share the objective of Equation~\ref{eq:agentic-routing-frontier} but differ in representation and training data. The zeroth policy $g^{(0)}$ is the open LightGBM ranker of Section~\ref{sec:singleton-routing}: hand-engineered features feed a shallow cost--quality ranker wrapped by the four cheap passes---order admission, demand construction, risk pricing, and capability matching~\citep{ke2017lightgbm}. It is deliberately not the final router; it is the cheapest policy that can run at every step and begin the arena-record stream of Section~\ref{sec:routing-data}. Every later policy $g^{(r)}_{\theta}$ is a learned model trained on that stream, and this subsection describes how such a policy is built, trained, and rolled forward.

\paragraph{Architecture.} A router-model iteration factorizes into three components. A \emph{state encoder} maps the harness state $h_t$---task and instruction, compressed and raw context, artifact and tool history, verifier outputs, recovery status, and recent route history---to a state vector, replacing the hand-engineered features of $g^{(0)}$. A \emph{supply encoder} maps each candidate model $m\in\mathcal{M}_t$ to a profile embedding built from its model card and observed outcome history: price, latency, context length, tool-use reliability, and task-family success. Because candidates are described by profiles rather than by fixed output classes, a newly released model can be scored from its profile without retraining a label space. A \emph{prediction head} then estimates, for a candidate action $u=(S,a)$ at state $h_t$, the expected task loss $\hat{\ell}(u\mid h_t)$ and expected cost $\hat{C}(u\mid h_t)$, and the router acts by exactly the rule of Equations~\ref{eq:singleton-clearing} and~\ref{eq:multimodel-clearing}, selecting $\arg\min_{u}[\hat{\ell}(u\mid h_t)+\lambda\hat{C}(u\mid h_t)]$.

\paragraph{Training.} Each iteration is trained to minimize the frontier objective of Equation~\ref{eq:agentic-routing-frontier} on the accumulated arena corpus $\mathcal{D}^{(r)}$ of Section~\ref{sec:routing-data}:
\begin{equation}
\theta^{(r+1)}=\arg\min_{\theta}\ \widehat{\mathbb{E}}_{\mathcal{D}^{(r)}}\big[\ell(\tau)+\lambda\,C(\tau)\big].
\label{eq:router-update}
\end{equation}
The expectation is an \emph{off-policy} estimate: because every record was produced by an earlier policy $g^{(r)}$ rather than by uniform sampling over actions, the estimator reweights logged outcomes with an inverse-propensity or doubly-robust correction, so the update targets outcome improvement rather than imitation of the logging policy~\citep{dudik2011doublyrobust}. No separate reward, recovery, or risk terms are needed: recovery cost and under-routing enter through the trajectory-level $C(\tau)$ and $\ell(\tau)$ exactly as in Sections~\ref{sec:singleton-routing} and~\ref{sec:multi-model-routing}, and $\lambda$ selects the operating point on the frontier.

\paragraph{Coverage.} Coverage is treated as a first-class part of the routing-data design. Because logged records naturally observe the chosen action most densely, the arena corpus includes the controlled exploration of Section~\ref{sec:routing-data}---uncertainty-triggered escalation, and occasional alternative-model or oracle replay on a sampled subset of states---to improve the counterfactual coverage that lets $g^{(r+1)}$ learn where $g^{(r)}$ under-routed a state or overpaid for capability.

\paragraph{Evolution.} The policies form a ladder, each generation promoted only if it moves the frontier of Equation~\ref{eq:agentic-routing-frontier} against the previous one. The released seed $g^{(0)}$ is the LightGBM ranker over hand-engineered features. The first learned generation $g^{(1)}$ replaces those features with a trained state encoder and prediction head, fitted offline on the first corpus and still gated behind the cheap admission filters of $g^{(0)}$. The next generation $g^{(2)}$ adds supply-encoder profiles for new-model generalization and exploration-filled counterfactuals, and can be distilled into a small fast router to stay within the step-level latency budget. Subsequent generations $g^{(r)}$ are continual updates as the corpus grows and the model pool changes. A candidate generation is shipped only when it lowers realized cost at equal task reward or raises task reward under a fixed budget; otherwise the deployment rolls back to the previous policy. Agreement with $g^{(r-1)}$ is never the target---movement of the quality--cost frontier is.

\paragraph{Deployment.} Because the router runs at every step, its own cost and latency are part of the trade-off. The deployment stays two-tier: the cheap admission filters and the $g^{(0)}$-style gate resolve clearly trivial or clearly hard states, and the heavier learned router is invoked only on the uncertain, high-value, or weakly recoverable states where a better matching decision is worth its overhead. This keeps the router on the useful side of its own cost--quality trade-off.

This staged path also fixes the open-source boundary. The $g^{(0)}$ seed can be released because it is inspectable, locally trainable, and useful for cold start; later generations depend on accumulated arena records, changing model-supply profiles, and continuous frontier evaluation, so they are an evolving \opensquilla policy path rather than a finalized artifact. Convergence of the router sequence is not required for deployment: each promoted generation is judged by measured frontier movement, and the arena-record stream supplies the data needed for continued policy improvement as the model pool and harness evolve.

\subsection{Routing Data as a Model-Arena Corpus}\label{sec:routing-data}

The data side of agentic routing is not an ordinary query log. A query log records what was asked and what a model answered. A harness-native routing record instead records a selected action and its later scoring by the execution environment. This distinction is central to the data flywheel: the router's decision is an action, while the outcome, verification result, recovery path, realized cost, and latency are labels supplied by the harness and runtime. The data is therefore a model-arena corpus rather than a transcript to be imitated.

For each decision point $t$, we store a typed arena record
\begin{equation}
\mathcal{R}_t
=
\Big(
q,\;
h_t,\;
\mathcal{M}_t,\;
\hat{\kappa}_t,\;
u_t,\;
s_t,\;
\mathrm{trace}_{t:T},\;
v_{t:T},\;
y,\;
C_{t:T},\;
D_{t:T},\;
\omega_t
\Big),
\end{equation}
where $q$ is the original task, $h_t$ is the harness state, $\mathcal{M}_t$ is the candidate model or tier slate available at the decision point, $\hat{\kappa}_t$ is the router's estimated capability demand, and
\begin{equation}
u_t=(S_t,a_t)
\end{equation}
is the routing action. The term $s_t$ stores the router-side scores, confidence, ranking margins, and exploration flags. The subsequent execution trace is $\mathrm{trace}_{t:T}$, the verification and runtime events are $v_{t:T}$, the final or intermediate outcome is $y$, the realized cost and duration are $C_{t:T}$ and $D_{t:T}$, and $\omega_t$ records provenance such as production exploitation, uncertainty-triggered exploration, offline replay, or oracle-style evaluation.

This schema deliberately separates prediction, action, and outcome. The fields $(h_t,\mathcal{M}_t,\hat{\kappa}_t,s_t)$ describe what the router believed before acting, the field $u_t$ records what it actually fielded, and the fields $(\mathrm{trace}_{t:T},v_{t:T},y,C_{t:T},D_{t:T})$ record how the environment scored that action. Because the action and its outcome are stored separately, a weak model choice is never the imitation target---the outcome is; we develop the consequences of this separation for self-poisoning in Section~\ref{sec:self-poison}.

The record is also the interface between the router and the rest of the harness. Routing-decision signals record the selected model or model set, optional harness configuration, candidate ranking, score margin, confidence, and aggregation policy. Tool-call signals record action schema, tool success, execution error, retry count, and recovery path. Context-management signals record context length, compression trigger, retained artifacts, dropped evidence, and post-compression verification. Verification and runtime signals record test results, judge outcomes, constraint violations, task-level success, realized cost, and latency overhead. These signals make the data harness-native: they describe not only what the model saw, but also how the execution system shaped, constrained, and evaluated the model's behavior.

These typed records accumulate into the arena corpus that drives the harness-native data flywheel of Section~\ref{sec:flywheel}. There we develop how the accumulated corpus is used to train routers, harness-native specialists, and harness policies (Section~\ref{sec:weak-supervision}); how controlled exploration---distinguished in the record by the provenance field $\omega_t$---maintains the coverage needed for off-policy learning without turning production into uncontrolled experimentation, and why the loop does not self-poison (Section~\ref{sec:self-poison})~\citep{dudik2011doublyrobust}; and why the accumulated corpus is hard to crawl, buy, or reconstruct (Section~\ref{sec:data-moat}). Here we fix only the data object: a typed, environment-labeled record of what was fielded and how the harness scored it.

\section{Experiments}\label{sec:experiments}

\subsection{Experimental Setup}\label{sec:exp-setup}

The experiments are organized around the two operating regimes of agentic routing. The first regime is singleton routing, where the router selects one model for the current turn or execution step and is evaluated by cost-quality trade-off. The second regime is multi-model ensemble routing, where the router selects several complementary models and an aggregator produces the final answer; this regime is evaluated by accuracy improvement under controlled additional cost. This separation is important because the two regimes answer different deployment questions: whether routing can make existing quality cheaper, and whether routing can make difficult tasks more accurate.

We evaluate on two main benchmarks. DRACO is a cross-domain deep-research benchmark that evaluates factual accuracy, completeness, objectivity, presentation quality, and citation quality across complex research tasks~\citep{zhong2026draco}. PinchBench (version~1.2.1) evaluates OpenClaw-style agents on real-world tasks and reports success rate, execution time, cost per task, and value score, making it suitable for cost-quality routing analysis. The present evaluation focuses on DRACO and PinchBench, which cover deep-research and OpenClaw-style agent execution.

We report benchmark-native quality, realized monetary cost, token usage, and latency, which together characterize the deployed cost--quality operating point of each policy. For singleton routing, these metrics test whether a mixed-model policy can preserve quality while reducing billed cost. For multi-model routing, they test whether complementary proposer sets and aggregation can move the frontier relative to strong singleton baselines. The same arena-record schema can support route-level analyses such as router accuracy, oracle gap, and aggregation success when replay or oracle data is available; the present evaluation focuses on end-to-end frontier movement.

\subsection{Singleton Routing: Cost--Quality Frontier Evaluation}\label{sec:singleton-exp}

This section evaluates the $K=1$ operating region of Equation~\ref{eq:agentic-routing-frontier}. Each row in Tables~\ref{tab:pinch-singleton} and~\ref{tab:draco-singleton} corresponds to a full benchmark execution of a deployed policy. The goal is to test whether singleton capability matching preserves aggregate task quality while reducing realized billed cost against fixed strong-model baselines. In these runs the router (Agentic routing) selects among Opus 4.8, GLM 5.2, and DS4 Flash on PinchBench, and among Opus 4.8, GLM 5.2, and DS4 Pro on DRACO; the strong-model baselines are Opus 4.8 run in the \opensquilla and OpenClaw harnesses.

\opensquilla uses singleton routing over the available model pool, while the OpenClaw baseline uses Claude Opus 4.8 as a fixed strong-model policy. We compare the routed \opensquilla policy against fixed strong-model policies under the same benchmark scoring protocol. The benchmark score measures end-to-end quality, while realized billed cost, token usage, and latency expose the operating point of each policy. This comparison follows the harness-centric evaluation view that the same base model can behave differently under different agent adapters, tool interfaces, and benchmark protocols~\citep{yang2024sweagent,zheng2026clawswebench}. The harness, benchmark split, and scoring protocol are held fixed within each aggregate setting.

\subsubsection{PinchBench}

\begin{table}[t]
\caption{Aggregate singleton-routing cost--quality operating points on PinchBench.}
\label{tab:pinch-singleton}
\small
\centering
\setlength{\tabcolsep}{6pt}
\renewcommand{\arraystretch}{1.12}
\begin{tabular*}{\textwidth}{@{}ll@{\extracolsep{\fill}}>{\columncolor{black!10}[\tabcolsep][2pt]}r>{\columncolor{black!10}[2pt][\tabcolsep]}rr@{}}
\toprule
Method & Agent & Score & Cost (\$) & Tokens (K) \\
\midrule
Opus 4.8 & OpenClaw & 93.35 & 0.2224 & 187.7 \\
OpenRouter Auto & OpenClaw & 88.10 & 0.1204 & 87.6 \\
Opus 4.8 & \opensquilla & 94.33 & 0.1649 & 97.7 \\
Agentic routing (ours) & \opensquilla & 93.14 & 0.0204 & 51.3 \\
\bottomrule
\end{tabular*}
\vspace{2pt}
\begin{minipage}{\textwidth}
\footnotesize\emph{Notes:} Score is the mean across tasks; costs are average billed cost per task; total tokens are input plus output tokens in thousands.
\end{minipage}
\end{table}

On PinchBench, the \opensquilla runs expose two singleton operating points relative to the fixed OpenClaw baseline, which scores $93.35$ at \$0.2224 per task. The routed configuration (Agentic routing) nearly matches this baseline while cutting cost by an order of magnitude: it scores $93.14$, a gap of only $0.21$ points and a quality retention of $99.77\%$, at \$0.0204 per task---a cost ratio of $9.17\%$ and roughly $10.9\times$ lower cost. The Opus4.8-backed \opensquilla point instead shifts toward quality: it scores $94.33$, about $0.98$ points above the OpenClaw baseline, while still reducing per-task cost to \$0.1649, a cost ratio of $74.1\%$. This second point provides a strong-model reference inside the \opensquilla harness and helps separate the cost-saving routed point from the quality-oriented fixed-model operating point. Table~\ref{tab:pinch-singleton} also includes OpenRouter Auto, OpenRouter's automatic model router (\url{https://openrouter.ai/docs/guides/routing/routers/auto-router}), as a routed baseline in the same OpenClaw harness: it scores $88.10$ at \$0.1204 per task, so our routed configuration is both $5.04$ points higher in quality and roughly $5.9\times$ cheaper than this query-level routing service.

\subsubsection{DRACO}

\begin{table}[t]
\caption{Aggregate singleton-routing cost--quality operating points on DRACO.}
\label{tab:draco-singleton}
\small
\centering
\setlength{\tabcolsep}{6pt}
\renewcommand{\arraystretch}{1.12}
\begin{tabular*}{\textwidth}{@{}ll@{\extracolsep{\fill}}>{\columncolor{black!10}[\tabcolsep][2pt]}r>{\columncolor{black!10}[2pt][\tabcolsep]}rr@{}}
\toprule
Method & Agent & Score & Cost (\$) & Tokens (K) \\
\midrule
Opus 4.8 & OpenClaw & 52.13 & 1.1420 & 54.2 \\
Opus 4.8 & \opensquilla & 52.36 & 0.6559 & 103.5 \\
Agentic routing (ours) & \opensquilla & 52.33 & 0.3729 & 108.6 \\
\bottomrule
\end{tabular*}
\vspace{2pt}
\begin{minipage}{\textwidth}
\footnotesize\emph{Notes:} Score is the mean across tasks; costs are average billed cost per task; total tokens are input plus output tokens in thousands; the routing threshold is set to $0.95$.
\end{minipage}
\end{table}

The DRACO aggregate shows a different savings mechanism. DRACO is particularly useful here because it scores complex deep-research outputs along factual accuracy, completeness, objectivity, presentation quality, and citation quality, rather than only exact-match task completion~\citep{zhong2026draco}. Agentic routing retains $99.94\%$ of the strong-model baseline score, moving from $52.36$ to $52.33$, while reducing per-task cost from \$0.6559 to \$0.3729. This corresponds to a cost ratio of $56.85\%$ and a cost reduction of $43.15\%$. Unlike PinchBench, the routed run uses slightly more input tokens than the fixed baseline. The cost reduction therefore cannot be explained only by prompt shortening or lower token volume. It indicates that singleton routing also changes the price composition of model calls: cheaper sufficient models can be fielded for parts of the execution while preserving aggregate quality.

\subsubsection{Summary}

Taken together, these aggregate frontier points support the singleton capability-matching view in Section~\ref{sec:singleton-routing}. The routed policy preserves the strong-agent operating region while reducing realized cost, suggesting that model capability can be allocated more selectively inside the harness without degrading end-to-end task quality. In PinchBench, the reduction comes from both lower token volume and cheaper model allocation. In DRACO, the routed policy preserves score while reducing cost despite using slightly more input tokens, indicating that the savings are not merely a byproduct of shorter context, but also reflect a different price composition of model calls across the execution trajectory.

The main takeaway is that singleton routing can shift the deployed agent to a more favorable cost--quality point. Rather than treating every execution state as requiring the strongest available model, the router fields capability according to the current state and the expected cost of failure or recovery. These aggregate runs therefore validate the practical feasibility of harness-native capability matching as a cost-saving operating mode. The next layer of analysis is route-level: associating each routing decision with the corresponding harness state, verifier signal, recovery event, and final task outcome. This finer-grained view will further explain the mechanism behind the observed frontier shift and guide later router-model iterations.

\subsection{Multi-Model Ensemble Routing: Accuracy and Cost--Quality Evaluation}\label{sec:multimodel-exp}

The multi-model-routing experiment tests whether selected complementary proposer sets and aggregation can improve the quality--cost frontier of agent execution. We compare fixed singleton baselines with selected multi-model routing configurations that vary proposer composition, sampling, and aggregation or prefiltering. The primary quantities are benchmark-native quality, average monetary cost per task, token or tool usage, and p50/p95 latency.

The multi-model experiments report benchmark-specific cost--quality operating points under fixed provider settings. Within each setting, we compare singleton baselines with a selected multi-model routing configuration, measuring benchmark-native quality, monetary cost, token usage, and latency. This evaluation characterizes whether complementary proposer sets and aggregation can move the deployed frontier relative to strong singleton execution.

DRACO is the primary benchmark for multi-model routing because deep research tasks often benefit from complementary model strengths: one model may retrieve better evidence, another may synthesize more coherent analysis, and another may produce more reliable citations. The aggregator can score candidate reports using factuality checks, citation coverage, completeness criteria, and consistency across model outputs. PinchBench provides short-task cost-quality evidence, and the Hermes mixture-of-agents configuration serves as an additional multi-model baseline in the DRACO comparison.

Unless otherwise noted, costs are averages per task, token counts are shown in thousands, and latencies are shown in seconds. Boldface highlights the primary comparison value in each table, usually the selected configuration's quality and, when applicable, its cost advantage. We use the benchmark's native score scale in each table; for example, PinchBench scores are normalized to $[0,1]$, and DRACO and Hermes report their native quality scores.

\subsubsection{DRACO}

For DRACO with the default DuckDuckGo web-search provider, Table~\ref{tab:draco-routing-results} reports singleton baselines and the selected multi-model routing configuration. The configuration uses DeepSeek V4, GLM 5.2, Gemini 3 Flash, and Qwen 3.7 as proposers, GLM 5.2 as the aggregator, and additional stochastic samples for the cheaper Gemini and Qwen proposers.

\begin{table}[t]
\caption{DRACO results comparing singleton baselines with the selected multi-model routing configuration.}
\label{tab:draco-routing-results}
\small
\centering
\setlength{\tabcolsep}{6pt}
\renewcommand{\arraystretch}{1.12}
\begin{tabularx}{\textwidth}{@{}>{\raggedright\arraybackslash}X>{\columncolor{black!10}[\tabcolsep][2pt]}r>{\columncolor{black!10}[2pt][\tabcolsep]}rrrrr@{}}
\toprule
\multirow{2}{*}{Method} & Quality & \multicolumn{4}{c}{Efficiency} & \multirow{2}{*}{Coverage} \\
\cmidrule(lr){2-2}\cmidrule(lr){3-6}
& Score & Cost (\$) & Tokens (K) & p50 (s) & p95 (s) & \\
\midrule
Fable 5 & 59.80 & 1.2122 & 93.7 & 187.7 & 343.5 & 94/100 \\
Opus 4.8 & 52.36 & 0.6559 & 103.5 & 165.3 & 270.6 & 100/100 \\
GPT-5.5 & 50.22 & 0.4505 & 81.9 & 278.1 & 457.4 & 100/100 \\
GLM 5.2 & 48.28 & 0.1214 & 116.8 & 207.0 & 351.5 & 100/100 \\
DeepSeek V4 Pro & 50.32 & 0.1320 & 83.4 & 213.1 & 346.0 & 100/100 \\
K2.7-code & 45.48 & 0.0676 & 86.3 & 186.2 & 491.7 & 100/100 \\
Qwen 3.7 & 49.34 & 0.0432 & 99.5 & 240.9 & 368.9 & 100/100 \\
Gemini 3 Flash & 40.79 & 0.0117 & 9.5 & 18.7 & 168.4 & 100/100 \\
\midrule
\textbf{Ours} & 60.82 & 0.3766 & 579.7 & 535.5 & 3097.0 & 100/100 \\
\bottomrule
\end{tabularx}
\vspace{2pt}
\begin{minipage}{\textwidth}
\footnotesize\emph{Notes:} Costs are averaged per task; token counts are shown in thousands and latencies in seconds. Values are rounded for display.
\end{minipage}
\end{table}

Among singleton baselines, Fable 5 is the strongest quality point, with an average score of $59.80$ and an average cost of \$1.2122 over the 94 tasks it executed. Our multi-model configuration reaches $60.82$ on the full 100-task DRACO run while reducing average cost to \$0.3766, improving quality by $1.02$ points and reducing cost by $68.9\%$ relative to Fable 5. This result illustrates the cost-efficient substitution regime predicted by the frontier formulation: complementary cheaper proposers plus an aggregator can outperform the strongest evaluated singleton baseline on both quality and monetary cost. As expected for an ensemble policy, this operating point spends additional wall-clock time to obtain a stronger quality--cost point; latency is an orthogonal deployment axis that can be controlled by parallel proposer execution, early stopping, or stricter proposer budgets.

Table~\ref{tab:draco-ensemble-comparison} directly compares our selected DRACO configuration with the Hermes MoA configuration. Hermes MoA is a preset-level virtual model provider: reference models first produce private advisory outputs, and the configured aggregator is the acting model that writes the assistant response and emits tool calls inside the normal Hermes loop~\citep{nousresearch2026hermesmoa}. Our configuration instead treats multi-model execution as a routed cost--quality operating point: the proposer set and aggregator are selected to move the benchmark frontier, so complementary cheaper models can substitute for a strong acting aggregator when the realized score justifies it. In this setting, our configuration improves the native score from $59.55$ to $60.82$ while reducing actual average cost from \$0.4460 to \$0.3766 per task, showing that the routing objective yields a better quality--cost point than the evaluated mixture-of-agents preset.

\begin{table}[t]
\caption{DRACO comparison of routing and baseline configurations.}
\label{tab:draco-ensemble-comparison}
\small
\centering
\setlength{\tabcolsep}{6pt}
\renewcommand{\arraystretch}{1.12}
\begin{tabularx}{\textwidth}{@{}>{\raggedright\arraybackslash}X>{\columncolor{black!10}[\tabcolsep][2pt]\raggedleft\arraybackslash}p{1.7cm}>{\columncolor{black!10}[2pt][\tabcolsep]\raggedleft\arraybackslash}p{1.7cm}@{}}
\toprule
Method & Score & Cost (\$) \\
\midrule
Hermes MoA & 59.55 & 0.4460 \\
Sakana/fugu-ultra & 46.70 & 0.6344 \\
\midrule
\textbf{Ours} & 60.82 & 0.3766 \\
\bottomrule
\end{tabularx}
\vspace{2pt}
\begin{minipage}{\textwidth}
\footnotesize\emph{Notes:} Costs use actual average cost per task for comparability. Hermes MoA denotes the evaluated Hermes mixture-of-agents configuration, with 100/100 coverage, 100 healthy tasks, no short outputs, and no judge errors. The sakana/fugu-ultra row uses DuckDuckGo search and 91 successful rows. Token usage is omitted because it is not reported for Hermes MoA.
\end{minipage}
\end{table}

We also evaluate the same DRACO setting with Brave as the web-search provider, replacing the default DuckDuckGo provider. Table~\ref{tab:draco-brave-routing-results} reports the singleton baselines and the selected multi-model configuration in this provider setting.

\begin{table}[t]
\caption{DRACO results with Brave web search comparing singleton baselines with the selected multi-model routing configuration.}
\label{tab:draco-brave-routing-results}
\small
\centering
\setlength{\tabcolsep}{6pt}
\renewcommand{\arraystretch}{1.12}
\begin{tabularx}{\textwidth}{@{}>{\raggedright\arraybackslash}X>{\columncolor{black!10}[\tabcolsep][2pt]}r>{\columncolor{black!10}[2pt][\tabcolsep]}rrrrr@{}}
\toprule
\multirow{2}{*}{Method} & Quality & \multicolumn{4}{c}{Efficiency} & \multirow{2}{*}{Coverage} \\
\cmidrule(lr){2-2}\cmidrule(lr){3-6}
& Score & Cost (\$) & Tokens (K) & p50 (s) & p95 (s) & \\
\midrule
Fable 5 & 62.06 & 1.3241 & 106.7 & 126.5 & 206.3 & 93/100 \\
Opus 4.8 & 59.11 & 1.6177 & 257.7 & 233.5 & 418.2 & 100/100 \\
GPT-5.5 & 53.28 & 0.8407 & 189.4 & 213.5 & 587.6 & 100/100 \\
\midrule
\textbf{Ours} & 64.09 & 0.1218 & 500.1 & 656.8 & 2096.0 & 100/100 \\
\bottomrule
\end{tabularx}
\vspace{2pt}
\begin{minipage}{\textwidth}
\footnotesize\emph{Notes:} Costs are averaged per task; token counts are computed as Avg Visible plus Avg Reason and shown in thousands; latencies are in seconds. All rows use Brave as the web-search provider. Fable 5 completed 93 of 100 tasks because safety filtering blocked 7 tasks, so its reported score, cost, token, and latency values are averaged over the 93 completed tasks.
\end{minipage}
\end{table}

With Brave search, Fable 5 is the strongest singleton quality point despite using less compute than Opus 4.8: over its 93 completed tasks, it scores $62.06$, $2.95$ points above Opus 4.8, while reducing average cost from \$1.6177 to \$1.3241 and token usage from $257.7$K to $106.7$K. The selected multi-model configuration improves DRACO quality by $2.03$ points over Fable 5, by $4.98$ points over Opus 4.8, and by $10.81$ points over GPT-5.5, while reducing average cost by $90.8\%$, $92.5\%$, and $85.5\%$, respectively. This is a stronger cost-efficient substitution point than the default-provider DRACO run: the ensemble benefits from complementary candidate generation while the aggregate monetary cost remains far below the singleton baselines. As expected for an ensemble policy, the configuration spends additional wall-clock time for the stronger quality--cost point, while deployment can control latency through parallel proposer execution, early stopping, or stricter proposer budgets.

\subsubsection{PinchBench}

For PinchBench, Table~\ref{tab:pinchbench-routing-results} reports the same compact metrics from the task-level evaluation workbook. We use the task rows rather than the workbook's overall summary row and report the selected multi-model routing configuration.

\begin{table}[t]
\caption{PinchBench results comparing singleton baselines with the selected multi-model routing configuration.}
\label{tab:pinchbench-routing-results}
\small
\centering
\setlength{\tabcolsep}{6pt}
\renewcommand{\arraystretch}{1.12}
\begin{tabularx}{\textwidth}{@{}>{\raggedright\arraybackslash}X>{\columncolor{black!10}[\tabcolsep][2pt]}r>{\columncolor{black!10}[2pt][\tabcolsep]}rrrr@{}}
\toprule
\multirow{2}{*}{Method} & Quality & \multicolumn{4}{c@{}}{Efficiency} \\
\cmidrule(lr){2-2}\cmidrule(l){3-6}
& Score & Cost (\$) & Tokens (K) & p50 (s) & p95 (s) \\
\midrule
Opus 4.8 & 0.9433 & 0.1649 & 97.7 & 23.1 & 150.1 \\
GPT-5.5 & 0.9373 & 0.0963 & 56.7 & 29.0 & 76.1 \\
\midrule
\textbf{Ours} & 0.9431 & 0.1349 & 272.4 & 96.0 & 223.7 \\
\bottomrule
\end{tabularx}
\vspace{2pt}
\begin{minipage}{\textwidth}
\footnotesize\emph{Notes:} Costs are averaged per task; token counts are input plus output tokens in thousands, and latencies are in seconds. PinchBench scores are normalized to $[0,1]$.
\end{minipage}
\end{table}

On PinchBench, the selected multi-model configuration essentially matches the Opus 4.8 quality point: the gap is only $0.0003$ score units, while average cost falls from \$0.1649 to \$0.1349, an $18.2\%$ reduction. Relative to GPT-5.5, the same configuration improves quality by $0.0058$ score units but uses higher cost and latency. On this saturated short-task benchmark, the relevant operating mode is cost-efficient quality parity: the selected multi-model configuration matches the strongest singleton quality point while reducing monetary cost.

\subsubsection{Routed Ensembles: Combining Agentic Routing with Aggregation}

The results above use \emph{selected} multi-model configurations: the proposer set is fixed in advance for each run. This subsubsection instead couples the agentic router with the aggregation stage: at each run the routing policy assembles the proposer set dynamically, and only the aggregator is held fixed. Using GLM 5.2 as the fixed aggregator, we evaluate three routing-policy operating points on DRACO with the default DuckDuckGo web-search provider: \emph{control}, the current routing policy; \emph{diversity-heavy}, which upweights the complementarity term of Equation~\ref{eq:multimodel-clearing} (a larger $\alpha$); and \emph{quality-heavy}, which upweights predicted per-model quality (a lighter cost weight $\lambda$). Table~\ref{tab:routed-ensemble-results} reports the three groups.

\begin{table}[t]
\caption{DRACO results for routed ensembles: the agentic router assembles the proposer set at run time under three routing policies, with GLM 5.2 held fixed as the aggregator.}
\label{tab:routed-ensemble-results}
\small
\centering
\setlength{\tabcolsep}{6pt}
\renewcommand{\arraystretch}{1.12}
\begin{tabularx}{\textwidth}{@{}>{\raggedright\arraybackslash}X>{\columncolor{black!10}[\tabcolsep][2pt]}r>{\columncolor{black!10}[2pt][\tabcolsep]}rrrrr@{}}
\toprule
\multirow{2}{*}{Routing policy} & Quality & \multicolumn{4}{c}{Efficiency} & \multirow{2}{*}{Coverage} \\
\cmidrule(lr){2-2}\cmidrule(lr){3-6}
& Score & Cost (\$) & Tokens (K) & p50 (s) & p95 (s) & \\
\midrule
Control & 59.18 & 0.3249 & 650.4 & 881.7 & 3122.6 & 100/100 \\
Diversity-heavy & 60.31 & 0.3172 & 586.6 & 837.1 & 2682.6 & 99/100 \\
Quality-heavy & 59.93 & 0.2582 & 721.3 & 1023.0 & 2439.0 & 100/100 \\
\bottomrule
\end{tabularx}
\vspace{2pt}
\begin{minipage}{\textwidth}
\footnotesize\emph{Notes:} Costs are averaged per judged task; token counts are total tokens divided by judged tasks, shown in thousands; latencies are in seconds. Coverage is judged tasks over attempted tasks.
\end{minipage}
\end{table}

Two observations follow. First, the diversity-heavy policy is the best routing policy---$60.31$ versus $59.18$ for control---at slightly lower cost and latency, which is direct evidence for the complementarity regularizer of Equation~\ref{eq:multimodel-clearing}: steering the router toward decorrelated proposer sets improves aggregate quality without additional spend. Second, the quality-heavy policy trades a small quality gap ($59.93$) for the lowest cost (\$0.2582), giving a cheaper operating point on the same frontier. The best routed group (diversity-heavy) reaches $60.31$ at \$0.3172, approaching the selected configuration of Table~\ref{tab:draco-routing-results} ($60.82$ at \$0.3766) while costing $15.8\%$ less and requiring no hand-picked proposer set.

\subsection{Cost--Score Frontier Summary}\label{sec:frontier-summary}

\begin{figure}[t]
\centering
\begin{tikzpicture}
\begin{axis}[
  width=0.5\linewidth, height=5.2cm,
  xmode=log, log basis x=10,
  xlabel={Cost per task (\$)}, ylabel={Score},
  title={(a) PinchBench}, title style={font=\small},
  xmin=0.014, xmax=0.36, ymin=92, ymax=95,
  xtick={0.02,0.05,0.1,0.2}, xticklabels={0.02,0.05,0.1,0.2},
  ytick={92,93,94,95},
  tick label style={font=\footnotesize}, label style={font=\small},
  grid=major, grid style={gray!18},
]
\addplot[oursblue, line width=1pt, mark=none] coordinates {(0.0204,93.14) (0.1349,94.31)};
\addplot[only marks, mark=o, mark size=2.6pt, draw=basegray, line width=0.9pt]
  coordinates {(0.1649,94.33) (0.0963,93.73)};
\addplot[only marks, mark=square*, mark size=3pt, color=oursblue] coordinates {(0.0204,93.14)};
\addplot[only marks, mark=star, mark size=4pt, color=oursblue] coordinates {(0.1349,94.31)};
\node[font=\scriptsize, anchor=west]  at (axis cs:0.0204,93.14) {\,Ours (single)};
\node[font=\scriptsize, anchor=north] at (axis cs:0.0963,93.73) {GPT-5.5};
\node[font=\scriptsize, anchor=east]  at (axis cs:0.1349,94.31) {Ours (multi)\,};
\node[font=\scriptsize, anchor=west]  at (axis cs:0.1649,94.33) {\,Opus 4.8};
\end{axis}
\end{tikzpicture}\hfill
\begin{tikzpicture}
\begin{axis}[
  width=0.5\linewidth, height=5.2cm,
  xmode=log, log basis x=10,
  xlabel={Cost per task (\$)}, ylabel={Score},
  title={(b) DRACO}, title style={font=\small},
  xmin=0.2, xmax=2, ymin=45, ymax=65,
  xtick={0.2,0.5,1,2}, xticklabels={0.2,0.5,1.0,2.0},
  ytick={45,50,55,60,65},
  tick label style={font=\footnotesize}, label style={font=\small},
  grid=major, grid style={gray!18},
]
\addplot[oursblue, line width=1pt, mark=none] coordinates {(0.3729,52.33) (0.3766,60.82)};
\addplot[only marks, mark=o, mark size=2.6pt, draw=basegray, line width=0.9pt]
  coordinates {(0.6559,52.36) (0.4505,50.22) (1.2122,59.80)};
\addplot[only marks, mark=square*, mark size=3pt, color=oursblue] coordinates {(0.3729,52.33)};
\addplot[only marks, mark=star, mark size=4pt, color=oursblue] coordinates {(0.3766,60.82)};
\node[font=\scriptsize, anchor=north] at (axis cs:0.3729,52.33) {Ours (single)};
\node[font=\scriptsize, anchor=south] at (axis cs:0.3766,60.82) {Ours (multi)};
\node[font=\scriptsize, anchor=north] at (axis cs:0.4505,50.22) {GPT-5.5};
\node[font=\scriptsize, anchor=west]  at (axis cs:0.6559,52.36) {\,Opus 4.8};
\node[font=\scriptsize, anchor=east]  at (axis cs:1.2122,59.80) {Fable 5\,};
\end{axis}
\end{tikzpicture}
\caption{Aggregate cost--score view on PinchBench and DRACO, combining the singleton and multi-model ensemble results. Gray circles are fixed singleton baselines; the blue square and star, joined by a line, are our singleton router and multi-model ensemble. Cost is shown on a log scale.}
\label{fig:cost-score-frontier}
\end{figure}
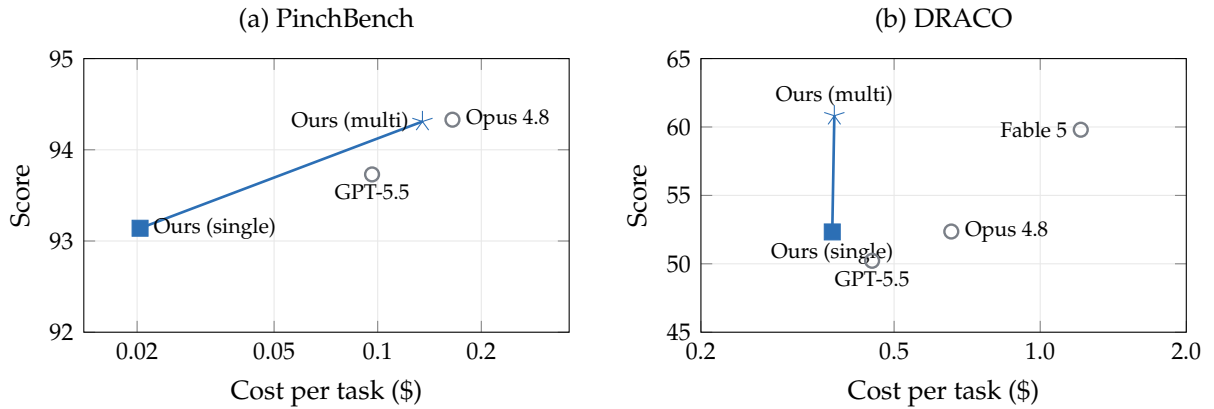

Figure~\ref{fig:cost-score-frontier} aggregates the singleton (Section~\ref{sec:singleton-exp}) and multi-model ensemble (Section~\ref{sec:multimodel-exp}) results into a single cost--score view for each benchmark, excluding the OpenClaw harness baseline and any baseline cheaper than our singleton router. Each point is a method, labeled by name; gray circles are fixed singleton baselines, and the blue square and star---joined by a line---are our singleton router and multi-model ensemble. On DRACO our two points dominate the strongest baselines: the multi-model ensemble exceeds both Opus 4.8 and Fable 5 in score at far lower cost, and the singleton router matches the Opus 4.8 quality point at a fraction of its cost. On PinchBench, where the strong baseline is already near saturation, the singleton router is the cheapest operating point and the ensemble nearly reaches the Opus 4.8 quality point at lower cost.

\section{The Harness-Native Data Flywheel}\label{sec:flywheel}

\subsection{Closed Loop}\label{sec:flywheel-loop}

The harness-native data flywheel describes how agent execution can generate the data needed to improve future agents. The loop is:
\begin{equation*}
\text{Base Models} \rightarrow \text{Harness Execution} \rightarrow \text{Routing Data} \rightarrow \text{Harness-Native Models} \rightarrow \text{Model Pool}.
\end{equation*}
Base models provide broad capabilities. The harness executes tasks and records routing decisions, traces, outcomes, and costs. The resulting data trains better routers and specialist models. These specialist models are added back into the model pool, where routing decides when they should replace or complement general foundation models. The important point is that the flywheel does not eliminate foundation models; it makes the model pool more heterogeneous and therefore makes routing more valuable.

The flywheel is budget-amplifying. Better routing increases the fraction of steps handled by cheap capable models. Lower average cost allows more tasks to be executed under the same budget. More tasks produce more trace data. More trace data improves routers and specialists, which further lowers cost or improves quality. The compounding effect is strongest early, when many routing and specialization opportunities remain, and gradually approaches a mature operating point as easy substitutions are exhausted and remaining tasks require strong models.

More precisely, the loop is an iteration on the frontier of Equation~\ref{eq:agentic-routing-frontier}. At iteration $r$, the routing policy $g^{(r)}$ executes inside the harness and appends its arena records to the corpus, $\mathcal{D}^{(r+1)}=\mathcal{D}^{(r)}\cup\mathcal{R}^{(r)}$; these records then update the router, the post-trained models, and the harness-native specialists. Progress is measured not by agreement with the previous policy but by \emph{frontier movement}: lower realized cost at comparable task reward, or higher task reward under a controlled budget.

\subsection{How to use the Accumulated Corpus}\label{sec:weak-supervision}

The value of the flywheel is not any single record but the accumulated corpus, which becomes a reusable training substrate for the whole system. As it grows, the same records are put to several uses. They train the router iterations of Section~\ref{sec:router-model-iterations}, improving demand construction, risk pricing, and capability matching. They expose repeated execution states---code repair, long-context recovery, tool-call formatting, or verification-aware editing---where a cheaper specialist could replace or complement a strong general model, yielding distillation and specialist-training sets. They supervise harness policies beyond model choice, such as when to compress context, when to verify, when to retry, and when to escalate. And they tell the operator what to build next: whether the next improvement should be a routing change, a distilled specialist, a verifier update, or a base-model post-training update.

The corpus also weakly supervises the models being routed. The routed action is weak evidence, not a gold label: the fact that the old policy selected a model does not make it optimal, and a skipped model is not necessarily wrong. Supervision comes from the environment's verdict---whether the output passed verification, whether it was later repaired, whether the task succeeded, and at what cost. A signal compiler $\Gamma$ maps each record into several post-training views,
\begin{equation}
\Gamma(\mathcal{R}_t)\rightarrow\{\mathcal{D}_{\mathrm{SFT}},\ \mathcal{D}_{\mathrm{pref}},\ \mathcal{D}_{\mathrm{reward}},\ \mathcal{D}_{\mathrm{distill}},\ \mathcal{D}_{\mathrm{curr}},\ \mathcal{D}_{\mathrm{safety}}\},
\end{equation}
so that a cheap-model failure that a strong model fixes becomes an SFT or distillation target, and a verified answer beating a failed one becomes a preference pair. The reward must be decomposed, because cost used directly as a language-model preference teaches brevity rather than genuine efficiency: quality and safety train the model, while cost and latency train the router,
\begin{equation}
R_{\mathrm{model}}(x,y)=Q(x,y)-\lambda_S S(x,y)-\lambda_E E(x,y),
\end{equation}
\begin{equation}
R_{\mathrm{route}}(x,y,m)=Q(x,y)-\lambda_C C(m,y)-\lambda_D D(m,y)-\lambda_{\mathrm{rec}} C_{\mathrm{rec}}(x,y,m).
\end{equation}
Across all of these uses the pattern is the same: routing first reveals a capability gap, and the accumulated corpus then indicates how to close it. This is what makes the corpus, rather than any single router, the durable asset of the flywheel.

\subsection{Why the Flywheel Does Not Self-Poison}\label{sec:self-poison}

A natural concern is that a router trained on data produced by earlier routers may simply imitate their mistakes. This concern applies to behavior cloning of the router's own decisions, but it does not describe the harness-native data schema. In our schema, the model choice is an action, while success, failure, recovery cost, and latency are labels supplied by the environment. A bad router that selects a weak model for a difficult state generates a negative example: this state-action pair led to failure or high recovery cost. The data is therefore logged feedback for off-policy learning, not a transcript to be copied blindly.

The schema is designed around five statistical requirements for reliable logged-feedback learning: coverage, selection-bias control, label variation, credit assignment, and verifier robustness. Coverage requires observing enough alternative actions for comparable states; selection-bias control accounts for the fact that logged records come from an earlier policy; label variation ensures that the corpus contains informative success and failure cases; credit assignment connects intermediate routing decisions to terminal outcomes inside long trajectories; and verifier robustness protects the learning signal when outcome labels are noisy. These requirements are addressed through uncertainty-triggered exploration, replay, and oracle traces; logged route probabilities and propensity-aware or doubly-robust estimation; a strong and diverse base-model pool; stored traces, intermediate verifier signals, and cost deltas; and multiple combined verification signals with audits of high-impact buckets.

\subsection{Why the Data Is Hard to Replace}\label{sec:data-moat}

Harness-native routing data is difficult to crawl or purchase because it is produced inside execution. Public text can show a final answer, and benchmark leaderboards can show aggregate scores, but neither contains the sequence of harness states, candidate model slates, route probabilities, rejected alternatives, tool errors, verification events, context-compression decisions, recovery costs, and exploration provenance that led to the outcome. This makes the data valuable even when parts of the router implementation are open-source. Table~\ref{tab:data-comparison} contrasts this harness-native arena corpus with ordinary web or preference data along the dimensions that make it hard to replace.

\begin{table}[t]
\caption{Harness-native routing data differs from ordinary web or preference data because it is produced by execution and labeled by the environment.}
\label{tab:data-comparison}
\small
\centering
\begin{tabularx}{\textwidth}{>{\raggedright\arraybackslash}p{0.25\textwidth}X X}
\toprule
Dimension & Ordinary interaction data & Harness-native arena corpus \\
\midrule
Reward signal & Often implicit, noisy, or preference-based. & Task success, verification result, recovery path, and exact cost. \\
Counterfactual value & Usually weak because unchosen model outputs are unavailable. & Strengthened by shadow evaluation, replay, and alternative-model audits. \\
Crawlability & Partly visible from public outputs. & Hidden inside runtime traces and harness state. \\
Structure & Usually weakly structured text or interaction logs. & Typed multi-field record with execution trace and state features. \\
Open-source implication & Code release may reveal much of the pipeline. & Code can be open while the arena corpus and high-accuracy router models remain proprietary. \\
\bottomrule
\end{tabularx}
\end{table}

\subsection{End State: A Three-Layer Model Pool}\label{sec:end-state}

The flywheel suggests a three-layer model pool. The first layer contains general foundation models that remain necessary for long-tail tasks, ambiguous user intent, and high-stakes reasoning. The second layer contains harness-native specialist models trained from execution traces for common task families, tool schemas, verification policies, or context-management regimes. The third layer contains ultra-cheap fast paths such as small models, classifiers, retrieval shortcuts, or cache hits for trivial decisions. Agentic routing is the policy that composes these layers into one execution system.

This end state changes the role of the harness. The harness is not merely an adapter that wraps a fixed model; it becomes the environment that defines tasks, supplies feedback, and cultivates specialist models. Model providers can continue to produce stronger general capabilities, while harness operators produce the execution data that determines how those capabilities should be deployed and distilled into specialists. The more diverse the model pool becomes, the more valuable routing becomes; the more mature the harness becomes, the better the training signal for future models. The more fine-grained the supply grows, the more valuable capability matching becomes, and the more precisely the corpus can indicate what to train next.

The flywheel has clear operating conditions. It is most valuable when the model pool remains heterogeneous, when routing overhead is kept below the savings from better allocation, and when environment labels are governed with provenance and multiple settlement signals. These conditions are also practical design constraints: cheap filters and LightGBM-style gates keep the router inexpensive, step-level routing keeps the decision frequency compatible with current harnesses, and typed arena records make verification and data governance explicit. Even when individual model capabilities converge, the execution traces retain independent worth for verification, recovery, and post-training because they record how the harness transforms model outputs into task outcomes.

\section{Conclusions and Future Work}\label{sec:conclusion}

This report argues that agentic routing is a first-class mechanism for agent systems because both sides of the agent equation---models and harnesses---are becoming specialized. The core idea is to route at execution steps using the full harness state, and to treat each routing decision as a source of structured supervision. This reframes cost control as a data problem: routing decisions create environment-labeled traces, traces train better routers and harness-native models, and those models return to the pool to improve future execution.

The reported experiments establish two deployed operating modes. In singleton routing, the open LightGBM-seeded policy preserves strong-agent quality while substantially lowering realized cost. In multi-model routing, complementary proposer sets and aggregation move the quality--cost operating point beyond strong singleton baselines on DRACO and match saturated PinchBench quality at lower cost. Together, these results support the central systems claim: routing inside the harness can allocate model capability more efficiently than fixed-model execution, while producing typed arena records for subsequent policy improvement.

Several extensions follow naturally from this framing. Edge-side routing raises a privacy--latency design point; cache-aware routing can preserve KV locality under model switching; automatic model profiling can refresh the supply encoder as new models arrive; and cross-harness transfer can test which state semantics generalize across execution substrates. Finally, finer-grained routing may eventually interact with speculative decoding and token-level model composition.

The broader conclusion is that harnesses do more than execute models. They define the environment in which model capabilities are observed, evaluated, and transformed into future training data. Agentic routing is the control layer that exposes this data stream. If agent-native training is the next stage of agent development, then harness-native routing is one concrete mechanism for producing the data that makes it possible.

\clearpage
\appendix
\section*{Appendix}
\section{Contributors}
Xinchen Liu, Hang Zhou, Yingjie Zong$^{\dagger}$, Yuchuan Tian$^{\dagger}$, Liuyang Song$^{\dagger}$, Shuo Zhang, Yulong Li, Wei He, Mengyu Zheng$^{\dagger}$, Runke Liu$^{\dagger}$, Siyang Cheng, Xiang Kuang, Hailin Hu, Kai Han$^{*}$, Yunhe Wang$^{*}$\blfootnote{$^{\dagger}$\,Intern. \quad $^{*}$\,Corresponding author.}

\section{Case Evidence for Single-Model Routing}
\label{app:router-case-evidence}

\subsection{Case A: CNC Platform Selection for Ti--6Al--4V Aerospace Machining}

\paragraph{Prompt (abridged).}
A precision machining shop in northern Mexico must choose between DMG MORI NLX 2500SY, Mazak Integrex i-400S, and Okuma Multus U4000 for Ti--6Al--4V aerospace components. The agent must compare spindle torque and rigidity, titanium tooling constraints including ceramic-tooling limitations, thermal compensation for $\pm$0.0005'' tolerance, Siemens NX CAM integration, AS9100D traceability through MTConnect/OPC-UA, Nuevo Le\'on service infrastructure, long-term operating costs, and the infeasible 15{,}000-hour annual utilization assumption. Under the high-confidence routing setting, the router selects DeepSeek V4 Pro rather than the fixed Opus baseline.

\footnotesize
\setlength{\tabcolsep}{3.0pt}
\renewcommand{\arraystretch}{1.13}

\begin{longtable}{L{0.13\linewidth}L{0.11\linewidth}L{0.44\linewidth}L{0.23\linewidth}}
\caption{Output evidence for Case A.}
\label{tab:case-cnc-router}\\
\toprule
\textbf{Method} & \textbf{Score / Cost} & \textbf{Logged output excerpt} & \textbf{Interpretation} \\
\midrule
\endfirsthead
\toprule
\textbf{Method} & \textbf{Score / Cost} & \textbf{Logged output excerpt} & \textbf{Interpretation} \\
\midrule
\endhead
\bottomrule
\endfoot

\rowcolor{black!6}Opus 4.8 &
52.42 \newline \$0.1663 &
``\textcolor{goodgreen}{15,000 annual hours is physically impossible--a calendar year contains only 8,760 hours}.'' The final answer gives a plausible high-level recommendation: NLX 2500SY for turning-heavy work, and Integrex i-400S or Multus U4000 for true 5-axis aerospace work. It also states, ``\textcolor{badred}{I was unable to retrieve live data--the search tool returned no results across many queries, and the DMG MORI, Mazak, and Okuma product pages were blocked}.'' &
Catches the infeasible utilization assumption and distinguishes machine classes, but leaves several specifications and the Nuevo Le\'on service comparison as items to verify. Evidence depth is weaker for traceability, service risk, and other rubric-weighted mechanisms. \\

\rowcolor{oursblue!10}Ours &
65.90 \newline \$0.1122 &
The answer is organized directly around the requested constraints: ``Spindle Torque \& Rigidity,'' ``Recommended Tooling for Ti--6Al--4V,'' ``Thermal Compensation for $\pm$0.0005 tolerance,'' ``Siemens NX CAM Integration,'' ``AS9100D Traceability via MTConnect / OPC-UA,'' ``Nuevo Le\'on Service Infrastructure,'' and ``The 15,000 Hours / Year Assumption.'' It explains that ceramics fail in titanium due to \textcolor{goodgreen}{chemical reactivity, low thermal conductivity, built-up edge, and thermal shock}. &
Returns a more complete requirement-by-requirement artifact, with broader coverage of tooling, thermal control, CAM integration, traceability, service risk, and utilization constraints. The extra task-specific evidence raises judged quality while the recorded cost remains lower. \\

\end{longtable}

\normalsize

\paragraph{Summary.}
The task's core difficulty is not simply recommending a CNC platform, but checking each machine against titanium machining constraints, CAM and traceability requirements, local service risk, and an infeasible utilization assumption. The fixed Opus 4.8 baseline catches the 15{,}000-hour issue but explicitly reports retrieval gaps and leaves several key comparisons as verification items. The routed DeepSeek V4 Pro trajectory returns a more complete requirement-level artifact, improving judged quality from 52.42 to 65.90 while reducing recorded cost from \$0.1663 to \$0.1122.

\subsection{Case B: AI Code-Completion Interface Timing and Trust}

\paragraph{Prompt (abridged).}
An enterprise software team is designing AI-powered code-completion interfaces. The agent must compare GitHub Copilot inline suggestions, Tabnine multi-line predictions, and Amazon CodeWhisperer comment-to-code generation across developers with 2--5 years versus 10+ years of experience. The prompt asks for evidence on suggestion latency thresholds, acceptance rates by task timing for debugging versus new feature development, explanation availability and trust calibration, Microsoft productivity studies, programmer-interruption research, and JetBrains AI assistant deployment metrics. Under the high-confidence routing setting, the router selects DeepSeek V4 Pro rather than the fixed Opus baseline.

\footnotesize
\setlength{\tabcolsep}{3.0pt}
\renewcommand{\arraystretch}{1.13}

\begin{longtable}{L{0.13\linewidth}L{0.11\linewidth}L{0.44\linewidth}L{0.23\linewidth}}
\caption{Output evidence for Case B.}
\label{tab:case-code-completion-router}\\
\toprule
\textbf{Method} & \textbf{Score / Cost} & \textbf{Logged output excerpt} & \textbf{Interpretation} \\
\midrule
\endfirsthead
\toprule
\textbf{Method} & \textbf{Score / Cost} & \textbf{Logged output excerpt} & \textbf{Interpretation} \\
\midrule
\endhead
\bottomrule
\endfoot

\rowcolor{black!6}Opus 4.8 &
36.60 \newline \$1.1753 &
``My live search backend returned empty results this session,'' so the answer says it has academic grounding but ``\textcolor{badred}{could not verify the proprietary industry numbers}'' requested for Microsoft productivity studies, JetBrains deployment metrics, and vendor latency data. It therefore treats millisecond thresholds as ``\textcolor{badred}{directional, not gospel}.'' &
Provides a useful conceptual frame, but the central product-design request needs concrete vendor and deployment evidence. The answer leaves several requested empirical metrics unverified, which is reflected in very low factual-accuracy scoring despite high cost. \\

\rowcolor{oursblue!10}Ours &
69.41 \newline \$0.1372 &
The answer builds a structured synthesis around programmer interruption costs, Copilot inline ghost text, Tabnine multi-line prediction, CodeWhisperer comment-to-code generation, experience-level stratification, latency bands, and explanation-driven trust calibration. It reports that debugging should favor on-demand help, while new-feature work can tolerate proactive completion, and highlights that \textcolor{goodgreen}{explanations improve trust calibration primarily by helping developers reject bad suggestions}. &
Returns a fuller design artifact: the output maps timing policy to task state, expertise level, suggestion format, and verification cost. It improves score from 36.60 to 69.41 while reducing cost by nearly an order of magnitude. \\

\end{longtable}

\normalsize

\paragraph{Summary.}
This task is difficult because it asks for a synthesis across HCI interruption research, product telemetry, vendor-specific interaction patterns, and design recommendations. The fixed Opus 4.8 run does useful academic framing but explicitly cannot verify several requested industry metrics. The routed DeepSeek V4 Pro trajectory produces a more actionable interface-design matrix and connects latency, task mode, seniority, and trust calibration to concrete product behavior, raising judged quality while lowering recorded cost from \$1.1753 to \$0.1372.

\subsection{Case C: Mumbai Podcast Studio Procurement}

\paragraph{Prompt (abridged).}
A professional podcast studio in Mumbai needs daily 90-minute, multi-guest production with 4--6 simultaneous participants. The agent must compare Rode RodeCaster Pro II, Zoom PodTrak P8, and Tascam Model 12 on dynamic-microphone preamp noise floor, Windows 11 USB stability for 2+ hour sessions, onboard processing versus post-production flexibility, daily-use durability under 80\%+ monsoon humidity, total system cost with cables and shock mounts, India-specific warranty coverage, and documented production-environment failure rates. Under the high-confidence routing setting, the router selects DeepSeek V4 Pro rather than the fixed Opus baseline.

\footnotesize
\setlength{\tabcolsep}{3.0pt}
\renewcommand{\arraystretch}{1.13}

\begin{longtable}{L{0.13\linewidth}L{0.11\linewidth}L{0.44\linewidth}L{0.23\linewidth}}
\caption{Output evidence for Case C.}
\label{tab:case-podcast-router}\\
\toprule
\textbf{Method} & \textbf{Score / Cost} & \textbf{Logged output excerpt} & \textbf{Interpretation} \\
\midrule
\endfirsthead
\toprule
\textbf{Method} & \textbf{Score / Cost} & \textbf{Logged output excerpt} & \textbf{Interpretation} \\
\midrule
\endhead
\bottomrule
\endfoot

\rowcolor{black!6}Opus 4.8 &
26.18 \newline \$0.8296 &
``My live search tools failed this session,'' and the answer says it retrieved ``only the TASCAM Model 12 official spec sheet and Rode's marketing page.'' It warns that \textcolor{badred}{prices, current warranty terms, and India availability must be re-verified} and refuses to fabricate precise failure-rate statistics. &
The answer is appropriately cautious, but the procurement task depends on concrete India-specific pricing, warranty, durability, and failure-risk comparisons. The resulting artifact remains incomplete for a buyer making a studio-standardization decision. \\

\rowcolor{oursblue!10}Ours &
61.67 \newline \$0.1828 &
The answer returns a comparison matrix covering preamp EIN / gain, Windows 11 USB stability, onboard processing, humidity durability, India street-price system cost, warranty and service, and production-environment failure risk. It also adds a Mumbai monsoon checklist: \textcolor{goodgreen}{dehumidifier, air conditioning during sessions, silica packs, dust cover, and voltage stabilizer / UPS}. &
Produces a more decision-ready procurement artifact: it turns the prompt into a total-cost and operational-risk comparison instead of a partial spec discussion. Quality rises from 26.18 to 61.67 while recorded cost falls from \$0.8296 to \$0.1828. \\

\end{longtable}

\normalsize

\paragraph{Summary.}
This task rewards procurement-specific completeness rather than generic product preference. The fixed Opus 4.8 baseline is transparent about missing live evidence but therefore leaves the buyer without the India-specific warranty, cost, and monsoon-risk details requested by the prompt. The routed DeepSeek V4 Pro trajectory supplies a structured operating-cost and deployment-risk comparison, improving judged quality by 35.49 points while reducing recorded cost by \$0.6468.

\normalsize

\section{Case Evidence for Multi-Model Ensemble Routing}
\label{app:ensemble-case-evidence}

This appendix expands the aggregate DRACO results with three same-prompt case studies selected from logged execution records. The selection rule is conservative: both strong singleton baselines must score below 60, while at least one routed ensemble trajectory must score above 70. Each subsection provides prompt context, the per-method score and realized cost, logged output excerpts, and a short interpretation of the failure mode. Scores are benchmark-native quality scores; reported costs are realized end-to-end USD charges, rounded to four decimals. In the per-case tables, each row is tinted by method---gray for Opus 4.8, amber for Fable 5, and blue for ours---and within each logged excerpt, \textcolor{goodgreen}{green} marks correct or concrete evidence while \textcolor{badred}{red} marks failures, omissions, or approximations.

\subsection{Case A: Fortive Segment-Level Financial Analysis}

\paragraph{Prompt (abridged).}
Analyze Fortive's evolving portfolio strategy after the Precision Technologies separation. The agent must calculate operating margins for Fortive's three segments using Q2 2025 revenue and operating profit, compare them with FY2024 segment margins, compute 2023--2024 revenue growth, and explain how the segment-level economics inform the portfolio strategy.

\footnotesize
\setlength{\tabcolsep}{3.0pt}
\renewcommand{\arraystretch}{1.13}

\begin{longtable}{L{0.13\linewidth}L{0.11\linewidth}L{0.44\linewidth}L{0.23\linewidth}}
\caption{Output evidence for Case A.}
\label{tab:case-fortive}\\
\toprule
\textbf{Method} & \textbf{Score / Cost} & \textbf{Logged output excerpt} & \textbf{Interpretation} \\
\midrule
\endfirsthead
\toprule
\textbf{Method} & \textbf{Score / Cost} & \textbf{Logged output excerpt} & \textbf{Interpretation} \\
\midrule
\endhead
\bottomrule
\endfoot

\rowcolor{black!6}Opus 4.8 &
32.28 \newline \$2.8554 &
``Data Gaps--What I Could NOT Verify: \textcolor{badred}{I was unable to retrieve the granular segment-level operating-profit tables} before reaching the research limit. Consequently, I \textcolor{badred}{cannot responsibly provide precise calculated operating margins} for each segment's Q2 2025 operating margin, FY2024 segment operating margins, or exact 2023 and 2024 segment revenue.'' &
Abstains from the central calculations, leaving the answer as qualitative strategy commentary. \\

\rowcolor{orange!12}Fable 5 &
34.81 \newline \$3.7695 &
``\textcolor{goodgreen}{I located and confirmed the primary sources}, but was \textcolor{badred}{cut off before extracting the exact segment tables}. The figures below are my \textcolor{badred}{best reconstruction}.'' The final table reports \textcolor{badred}{approximate margins}, including IOS at 27--28\%, Precision Technologies at 23--24\%, and Advanced Healthcare Solutions at 14--15\% for Q2 2025. &
Finds the right source family but substitutes approximate reconstructed ranges for required numerator/denominator calculations. \\

\rowcolor{oursblue!10}Ours &
72.15 \newline \$2.1328 &
``Three-Segment View'': IOS revenue \$675.7M, operating profit \$166.7M, \textcolor{goodgreen}{margin 24.7\%}; Precision Technologies revenue \$523.6M, operating profit \$93.2M, \textcolor{goodgreen}{margin 17.8\%}; Advanced Healthcare Solutions revenue \$319.5M, operating profit \$39.1M, \textcolor{goodgreen}{margin 12.2\%}. The same answer reports FY2024 margins of 26.0\%, 22.4\%, and 12.1\%, and 2023--2024 revenue growth of 3.9\%, 0.3\%, and 4.7\%. &
Returns a reproducible segment table and the cross-period comparison needed for the strategic conclusion. \\

\end{longtable}

\normalsize

\paragraph{Summary.}
The task's core difficulty is not fluent financial commentary but exact extraction of segment-level numerators and denominators. Opus 4.8 explicitly abstains from the requested calculations, while Fable 5 substitutes approximate reconstructed ranges for the required values. The routed ensemble returns a reproducible calculation table and the cross-period comparison needed for the strategic conclusion. This case illustrates how complementary retrieval paths and aggregation reduce rubric-weighted factual omissions while also lowering cost relative to both singleton baselines.

\subsection{Case B: Cross-Cultural Checkout Error Prevention and Recovery}

\paragraph{Prompt (abridged).}
Research how error prevention and recovery mechanisms in e-commerce checkout flows accommodate different cultural attitudes toward mistakes and system trust across Japan, Germany, and Brazil. The agent must compare confirmation dialogs, undo capabilities, and error-message tone in Rakuten, OTTO, and Mercado Livre, using uncertainty-avoidance and high-/low-context communication as the synthesis frame.

\footnotesize

\begin{longtable}{L{0.13\linewidth}L{0.11\linewidth}L{0.44\linewidth}L{0.23\linewidth}}
\caption{Output evidence for Case B.}
\label{tab:case-checkout}\\
\toprule
\textbf{Method} & \textbf{Score / Cost} & \textbf{Logged output excerpt} & \textbf{Interpretation} \\
\midrule
\endfirsthead
\toprule
\textbf{Method} & \textbf{Score / Cost} & \textbf{Logged output excerpt} & \textbf{Interpretation} \\
\midrule
\endhead
\bottomrule
\endfoot

\rowcolor{black!6}Opus 4.8 &
50.39 \newline \$1.7531 &
``Rakuten Ichiba: Japanese e-commerce is known for \textcolor{badred}{information-dense layouts, long product pages, and multiple review/confirmation stages}.'' The answer describes \textcolor{badred}{Germany as relying on law and payment timing, and Brazil as relying on Mercado Pago buyer protection}. &
Captures the high-level cultural frame but omits platform-level cancellation controls, button wording, and refund timing. \\

\rowcolor{orange!12}Fable 5 &
55.92 \newline \$1.9201 &
``\textcolor{badred}{My research was cut short by tool failures}.'' The answer identifies a general Rakuten order confirmation screen, \textcolor{goodgreen}{the German final purchase-button rule, OTTO's 30-day return window, and Mercado Pago escrow}, but \textcolor{badred}{still omits Rakuten's final confirmation label, Rakuten's time-boxed cancellation window, and Mercado Livre's explicit cancellation capability}. &
Improves specificity but still misses several recovery controls emphasized by the rubric. \\

\rowcolor{oursblue!10}Ours &
87.38 \newline \$0.2563 &
``Communication Context: Japan--high-context; Germany--low-context; Brazil--high-context.'' The answer then states that Rakuten uses \textcolor{goodgreen}{a final ``Confirm Order'' control and a cancellation window that can be as narrow as 15--30 minutes}; OTTO is shaped by the Button Solution and self-service Storno cancellation; and Mercado Livre supports ``Cancelar compra'', Compra Garantida, \textcolor{goodgreen}{a seven-day right of withdrawal}, and Mercado Pago-mediated recovery. &
Combines the cultural frame with concrete platform mechanisms, making the UX synthesis actionable. \\

\end{longtable}

\normalsize

\paragraph{Summary.}
Both singleton baselines capture part of the high-level cultural story, but they under-specify the platform mechanisms that make the UX analysis actionable. Opus 4.8 remains mostly at the abstraction level of interface explicitness, legal recourse, and escrow protection. Fable 5 is more specific, but still misses several rubric-weighted controls and recovery paths. The routed ensemble combines cultural framing with platform evidence and therefore produces a stronger synthesis at 13--15\% of the singleton baseline costs.

\subsection{Case C: Coffee Smallholder Traceability Platform Selection}

\paragraph{Prompt (abridged).}
A cooperative of 5,000 smallholder coffee farmers in Colombia, Ethiopia, and Indonesia needs to choose a farm management and traceability platform. The agent must compare Olam AtSource, FairTrace, and alternatives on basic-phone usability, mountainous GPS plot mapping, Rainforest Alliance and Fairtrade certification requirements, multilingual support, offline functionality, and cost for a limited-budget cooperative.

\footnotesize

\begin{longtable}{L{0.13\linewidth}L{0.11\linewidth}L{0.44\linewidth}L{0.23\linewidth}}
\caption{Output evidence for Case C.}
\label{tab:case-coffee}\\
\toprule
\textbf{Method} & \textbf{Score / Cost} & \textbf{Logged output excerpt} & \textbf{Interpretation} \\
\midrule
\endfirsthead
\toprule
\textbf{Method} & \textbf{Score / Cost} & \textbf{Logged output excerpt} & \textbf{Interpretation} \\
\midrule
\endhead
\bottomrule
\endfoot

\rowcolor{black!6}Opus 4.8 &
49.11 \newline \$1.5008 &
``FairTrace is ambiguous. \textcolor{badred}{I could not verify a distinct, established product by that exact name} in coffee traceability. The closest real tools are Fairfood's Trace.'' The answer \textcolor{goodgreen}{correctly describes an offline field-agent model} and notes that Rainforest Alliance data must feed into MultiTrace. &
Recognizes the farm-tool/certification distinction in part, but misreads FairTrace and omits several relevant first-mile platforms. \\

\rowcolor{orange!12}Fable 5 &
55.75 \newline \$0.7898 &
``FairTrace--\textcolor{badred}{I could not verify an active, established platform under exactly this name}. You may be thinking of Fairfood's Trace.'' The answer lists \textcolor{goodgreen}{Farmforce, KoltiTrace, SourceTrace, Cropin, and Fairfood Trace}, and notes offline-first capture and GPS drift. &
Broadens vendor coverage but still treats FairTrace as a vendor-name ambiguity rather than a mandatory Fairtrade transaction and assurance portal. \\

\rowcolor{oursblue!10}Ours &
83.63 \newline \$0.0322 &
``\textcolor{goodgreen}{FairTrace is best understood as a certification/compliance ledger for Fairtrade certified cooperatives}: it records transactions, premiums, and assurance evidence. \textcolor{goodgreen}{It is not a farm management information system}.'' The answer recommends a \textcolor{goodgreen}{two-tier architecture}: mandatory certification portals such as FairTrace and RA MultiTrace, plus first-mile field apps such as Farmforce, KoltiTrace, SourceTrace, Cropin, or Farmerline-style tools for offline registration, GPS mapping, purchase records, and multilingual farmer communication. &
Resolves the system boundary first, then recommends a layered deployment architecture rather than a flat vendor comparison. \\

\end{longtable}

\normalsize

\paragraph{Summary.}
The main error in the singleton baselines is a taxonomy error: certification portals and first-mile field applications are treated as comparable products. Opus 4.8 recognizes the need for field-agent workflows but fails to identify several relevant platforms and misreads FairTrace. Fable 5 broadens vendor coverage but still does not state the mandatory role of certification portals or their complementarity with field apps. The routed ensemble resolves the system boundary first and then recommends a layered architecture, yielding a large quality gain at a small fraction of the baseline costs.

\bibliography{sn-bibliography}

\end{CJK*}
\end{document}